\documentclass[journal]{IEEEtran}
\usepackage{amsmath,amsfonts}
\usepackage{algorithm}
\usepackage[caption=false,font=normalsize,labelfont=sf,textfont=sf]{subfig}
\usepackage{url}
\usepackage{cite}
\usepackage{algorithmicx}
\usepackage{algpseudocode}
\usepackage{multirow}

\newcommand{\etal}{\textit{et al.}}
\usepackage{microtype}      
\usepackage[table]{xcolor}
\usepackage{bm}
\usepackage{booktabs}
\usepackage{adjustbox}

\begin{document}

\title{Unlearnable Examples Give a False Sense of Data Privacy: Understanding and Relearning}

\author{Pucheng Dang,
        Xing Hu$^{*}$~\IEEEmembership{Member, IEEE},
        Kaidi Xu~\IEEEmembership{Member, IEEE},
        Jinhao Duan, \\
        Di Huang,
        Husheng Han,
        Rui Zhang,
        Zidong Du~\IEEEmembership{Member, IEEE},

\thanks{Pucheng Dang and Husheng Han are with the SKL of Processors, Institute of Computing Technology, CAS, Beijing 100190, China, the University of Chinese Academy of Sciences, Beijing, 100049, China.
E-mail:\{dangpucheng20g,hanhusheng20z\}@ict.ac.cn
}
\thanks{Xing Hu and Zidong Du are with the SKL of Processors, Institute of Computing Technology, CAS, Beijing, 100190, China, Shanghai Innovation Center for Processor Technologies, Shanghai, 201210, China.
E-mail:\{huxing,duzidong\}@ict.ac.cn
}
\thanks{Kaidi Xu and Jinhao Duan are with the Department of Computer Science, College of Computing \& Informatics, Drexel University. Philadelphia, 19104, USA.
E-mail:\{kx46,jd3734\}@drexel.edu
} 
\thanks{Huang Di and Rui Zhang are with the SKL of Processors, Institute of Computing Technology, CAS, Beijing, 100190, China.
E-mail:\{huangdi,zhangrui\}@ict.ac.cn
} 
\thanks{$^{*}$Corresponding authors: Xing Hu}
}

\markboth{Journal of \LaTeX\ Class Files,~Vol.~14, No.~8, August~2021}%
{Shell \MakeLowercase{\textit{et al.}}: A Sample Article Using IEEEtran.cls for IEEE Journals}


\maketitle

\begin{abstract}
Unlearnable examples are proposed to prevent third parties from exploiting unauthorized data, which generates unlearnable examples by adding imperceptible perturbations to public publishing data. These unlearnable examples proficiently misdirect the model training process, leading it to focus on learning perturbation features while neglecting the semantic features of the image. 
In this paper, we make an in-depth analysis and observe that models can learn both image features and perturbation features of unlearnable examples at an early training stage, but are rapidly trapped in perturbation features learning since the shallow layers tend to learn on perturbation features and propagate harmful activations to deeper layers. 
Based on the observations, we propose \textit{Progressive Staged Training}, a self-adaptive training framework specially designed to break unlearnable examples. The proposed framework effectively prevents models from becoming trapped in learning perturbation features. 
We evaluated our method on multiple model architectures over diverse datasets, e.g., CIFAR-10, CIFAR-100, and ImageNet-mini. Our method circumvents the unlearnability of all state-of-the-art methods in the literature, revealing that existing unlearnable examples give a false sense of privacy protection and provide a reliable baseline for further evaluation of unlearnable techniques.

\end{abstract}


\section{Introduction}
\IEEEPARstart{D}{eep} neural networks (DNNs) have significantly boosted computer vision techniques in the past decade and achieved even better capabilities surpassing human beings~\cite{resnet,vgg16-bn,densenet121,gan,mrcnn,vit,swim-transformer,diffusion,clip,DiffusionDet,tcsvt1,tcsvt2,tcsvt3,tcsvt4,tcsvt5}. One of the most important factors contributing to the great success comes from the representative and valuable data collected for training. Not only influential public datasets such as CIFAR~\cite{cifar} and ImageNet~\cite{imagenet} are proposed for benchmark evaluation, but also huge volumes of scenario-sensitive data are used for real model training and deployment~\cite{1-1-6, 1-1-5,datasetadd1,datasetadd2}. Data has become an important asset and ensuring authorized access to sensitive data is crucial. 

\begin{figure}[t]
\vskip -0.1in
\begin{center}
\centerline{\includegraphics[width=\columnwidth,height=0.6\columnwidth]{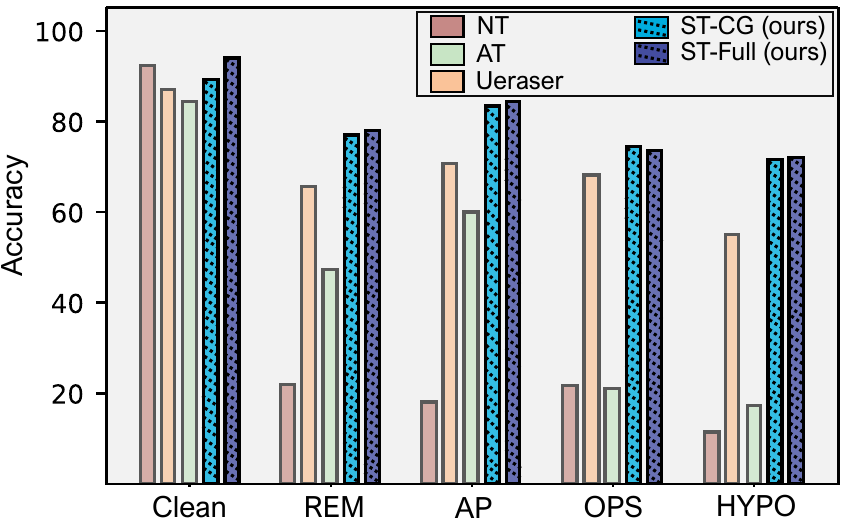}}
\vskip -0.1in
\caption{\footnotesize{Test accuracy of a ResNet-18 with normal (NT), adversarial 
 (AT), Chroma-CVE and ST series trained (ours) on various unlearnable protected CIFAR-10 datasets such as REM~\cite{REM}, SP~\cite{SP} OPS~\cite{OPS} and HYPO~\cite{HYPO}. 
 The results reflect that our training method ST defeats the data protection capability of these unlearnable methods and is highly effective in preserving clean accuracy.}}
\label{bar}
\end{center}
\vskip -0.25in
\end{figure}

\begin{figure*}[t]
\begin{center}
	\includegraphics[width=\linewidth]{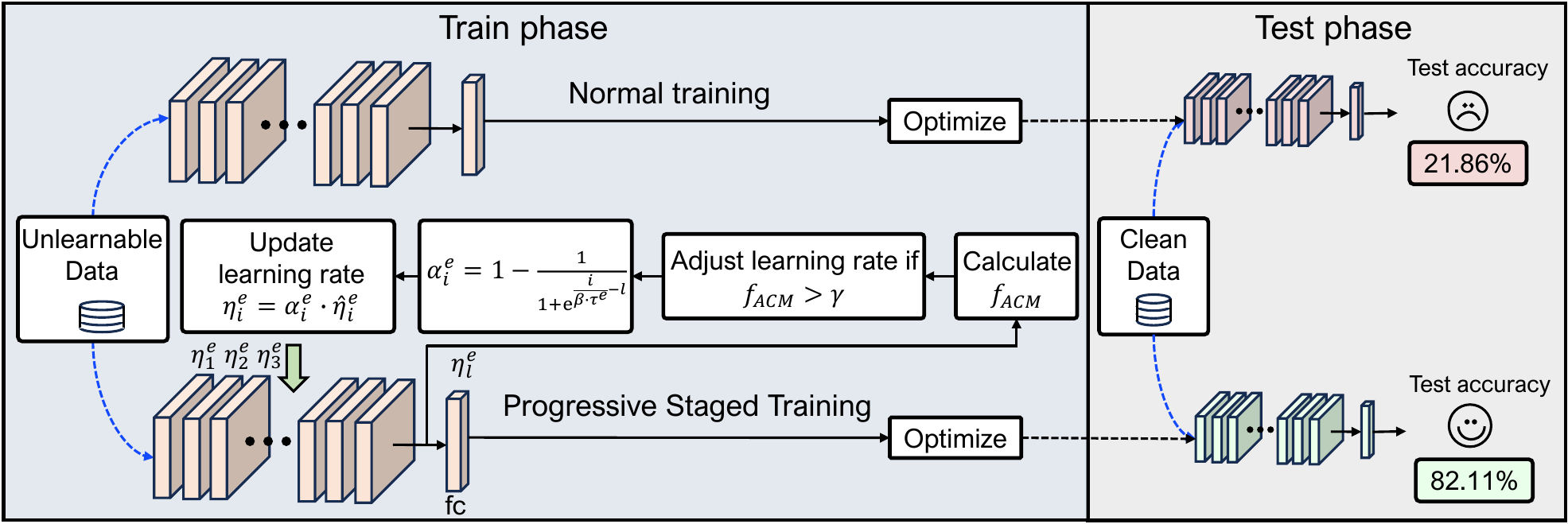}
    \vspace{-0.2in}
\caption{\footnotesize{An overview of ST when countering unlearning perturbations. Compared with normal training, once the indicator \textit{ACM} (Eq.~\eqref{eq:acm}) detects unlearnable perturbation feature learning, progressive staged training (ST) recovers clean accuracy of unlearnable examples by adjusting the learning rate of each layer (Eq.~\eqref{eta}). }}
\label{overview}
\end{center}
\vskip -0.25in
\end{figure*}

\subsection{Current Research Status of Unlearnable Example}
Previous studies propose ``unlearnable example'' techniques to address unauthorized access to sensitive data~\cite{EM, REM, SP, TUE, OPS, UC}. Specially, Yu \etal~\cite{SP} found that perturbations act as shortcuts~\cite{shortcutlearning1,shortcutlearning2,shortcutlearning3,shortcutlearning4,shortcutlearning5,shortcutlearning6} to mislead models to a bad state. These unlearnable examples are generated by injecting carefully crafted, imperceptible perturbations into the training data, which causes models to focus on learning irrelevant perturbation features rather than the meaningful semantic features of natural images.
Meanwhile, some data poisoning attacks~\cite{2-2-1,AR,HYPO,poisonattack1,poisonattack2} aim to manipulate the performance of a machine learning model via injecting maliciously poisoned examples into its training set, and such attacks can also be employed for data protection purposes. 

To address the challenge of learning from unlearnable samples, several approaches have been proposed~\cite{ueraser,LE,avatar,newadd2}. Ueraser~\cite{ueraser} employs data augmentation techniques to reduce the impact of unlearnable perturbations. LE~\cite{LE} and AVATAR~\cite{avatar} utilize popular diffusion models to filter out these unlearnable perturbations. However, these approaches primarily focus on data processing methods for handling unlearnable samples, rather than investigating how unlearnable perturbations function from a model learning perspective or exploring ways to mitigate the influence of such perturbations through improvements in the model's learning process.

\subsection{Motivations and Observations}
To address this research gap, we investigate the vulnerabilities of unlearnable examples from a model learning perspective and design methods to neutralize them by adjusting the model's learning process.
We initially examine the differential behaviors exhibited by the model when presented with unlearnable examples versus clean examples.
Specifically, we make the following observations: 1) In the early stages of training, models learn both unlearnable perturbation features and image semantic features. However, in subsequent stages, the focus shifts exclusively to the acquisition of unlearnable perturbation features.  This observation reveals that unlearnable examples are indeed capable of being learned: by guiding the model to focus on image semantic features after the initial training phase, it is possible to extract semantic knowledge from samples initially deemed unlearnable. 
2) Compared to deeper layers, correctly guiding the model's shallow layers to learn image semantic features plays a crucial role in preventing the model from becoming trapped in learning unlearnable perturbation features. When shallow layers are trapped in learning unlearnable perturbation features, detrimental activation is propagated through these layers, contaminating deeper layers and increasing their propensity for wrong learning. This observation indicates that the quality of shallow-layer training is significantly correlated with the overall tendency of the model to break through unlearnable perturbation learning.

\subsection{Contributions}
Based on these observations, we provide the following suggestions: 1) Models can extract valid semantic features from unlearnable examples during the initial phase of training. 2) Correctly guiding the model's shallow layers to learn image semantic features is the key to defeating unlearnable examples. 
In this case, we propose a training paradigm called Progressive Staged Training (ST). ST aims to prevent models from becoming trapped in unlearnable perturbation feature learning. ST assesses whether the models are in the wrong learning process and implements a progressive, staged strategy for training that involves dynamically adjusting the learning rate.
To quantitatively evaluate whether the model is trapped in the wrong learning process, we propose Activation Cluster Measurement (\textit{ACM}), based on the insight that unlearnable perturbation features have a larger inter-class distance and smaller intra-class distance~\cite{SP} in a low dimension space. To demonstrate the efficacy of ST, we conduct comprehensive experiments with multiple model architectures on CIFAR-10~\cite{cifar}, CIFAR-100~\cite{cifar}, and ImageNet-mini. Experimental results show that ST significantly defeats existing unlearnable techniques, thereby providing a dependable benchmark for the subsequent evaluation of samples considered unlearnable.

The scope of this paper is to invalidate unlearnable examples, and our method is a special design for unlearnable examples. 
As shown in Fig.~\ref{bar}, our method prevents models from the trap caused by unlearnable perturbation. 
Furthermore, it has proven to be highly efficient in maintaining accuracy with clean data.
We summarize our contributions as follows: 

\begin{itemize}
    \item  We demonstrate how models can learn valid image semantic features from initially unlearnable examples during the early stages of training. Guiding shallow layers to learn image semantic features substantially aids in defeating unlearnable examples.
    \item  We propose a learning framework, ST, to prevent models from being trapped in unlearnable perturbation feature learning, with the help of \textit{ACM} metric for quantitatively evaluating whether the model is in the wrong learning process.
    \item  We perform comprehensive experiments to show that our ST framework breaks all state-of-the-art unlearning samples in the literature on multiple model architectures across various datasets.
\end{itemize}

\section{Related work}

Unlearnable examples~\cite{EM,REM,SP,TUE,OPS,UC} share similar mechanisms as data poisoning attacks~\cite{2-2-1,AR,HYPO,poisonattack1,poisonattack2} but serve different application scenarios. The unlearnable example is a data protection method that adds imperceptible perturbations to the entire dataset. Suppose we have a dataset consisting of original clean examples $\mathcal{D}_{clean}=\{\left(\bm{x}_i,{y}_i\right)\}_{i=1}^{n}$ drawn from a distribution $\mathcal{S}$, where $\bm{x}_i\in \mathcal{X}$ is an input image and ${y}_i\in\mathcal{Y}$ is the associated label. We assume that the unauthorized parties will use the published training dataset to train a classifier $f_{\bm{\theta}}:\mathcal{X}\rightarrow\mathcal{Y}$ with parameter $\bm{\theta}$. Error-minimizing (EM)~\cite{EM} perturbations are produced by an alternating bi-level min-min optimization on both model parameters and perturbations:
\begin{equation}
    \arg\min_{\bm{\theta}}\mathbb{E}_{(\bm{x}_i,{y}_i)}\left[\min_{\bm{\delta}i}\mathcal{L}\left(f_{\bm{\theta}}(\bm{x}_i+\bm{\delta}_i),y_i\right)\right]
\end{equation}
where $\mathcal{L}$ is the cross entropy loss and $\Vert\bm{\delta}_{i}\Vert_{\infty}<\epsilon$. Being induced to trust that the perturbation can minimize the loss better than the original image features, the model will pay more attention to the perturbations. However, EM perturbations have no defense on adversarial training which adds random perturbations to input while training to improve the robustness of models. To solve this problem robust error-minimizing (REM)~\cite{REM} perturbations implement adversarial training while generating unlearnable perturbations, which grants perturbations the ability to resist adversarial training. 
Yu, \etal found that perturbations act as shortcuts to mislead models to a bad state, and proposed synthetic perturbations (SP)~\cite{SP} which generate unlearnable perturbations without the optimization of error-minimization. One-pixel shortcut (OPS)~\cite{OPS} used a perceptible pixel as the perturbation to protect data. To improve the transferability of unlearnable perturbations, Ren, \etal proposed transferable unlearnable examples~\cite{TUE}, which protects data privacy from unsupervised learning. However, such unlearnable examples are a special design for unsupervised learning and do not perform well with adversarial training. Unlearnable Clusters~\cite{UC} provides unlearnable examples for label-agnostic learning in which attackers could define labels by themselves. 

Meanwhile, some data poisoning attacks~\cite{2-2-1,AR,HYPO} aim to manipulate the performance of a machine learning
model via injecting maliciously poisoned examples into its training set and such attacks can also be employed for data protection purposes. 
Adversarial Examples as Poisons (AEP)~\cite{2-2-1} show that adversarial examples with the original labels are strong poisons at training time.
Hypocritical perturbations (HYPO)~\cite{HYPO} considers the adversarial sample generation and uses a pre-trained surrogate to generate adversarial samples for unlearnability. 
Autoregressive poisoning (AP)~\cite{AR} proposes a generic perturbation that can be applied to different datasets and architectures. 

Recent works have explored learning from unlearnable samples. Ueraser~\cite{ueraser} employs data augmentation techniques to mitigate the impact of unlearnable perturbations. LE~\cite{LE} and AVATAR~\cite{avatar} utilize popular diffusion models to filter out these unlearnable perturbations. These approaches primarily focus on data processing methods for handling unlearnable samples, whereas our work is orthogonal, aiming to optimize the learning process by dynamically adjusting the learning rate to effectively learn from such samples.

\begin{figure}[t]
\begin{center}
\centering
    \includegraphics[width=0.89\linewidth]{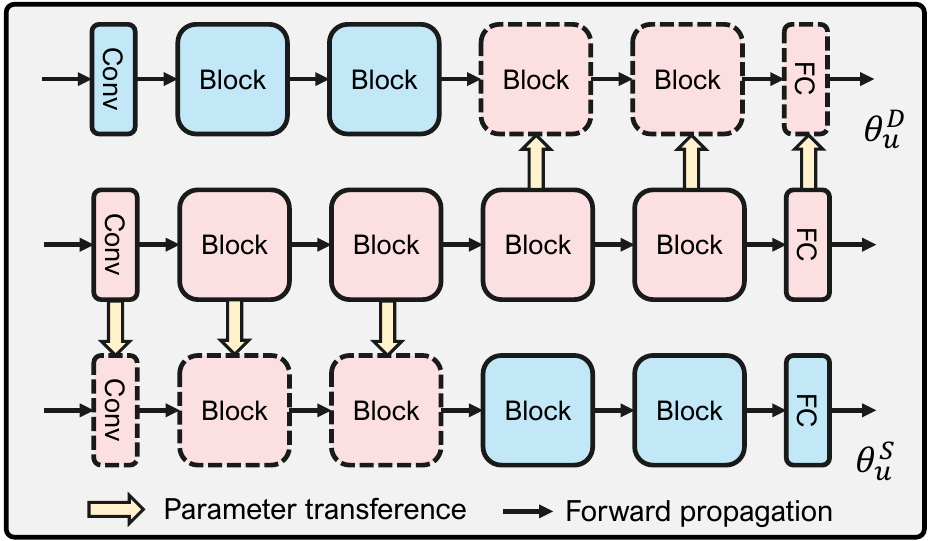}
    \caption{\footnotesize{The sketch of $\bm{\theta}_u^D$ and $\bm{\theta}_u^S$. Their weights come from a model trained on clean data (middle row) by replacing deep or shallow layers. The result of $\bm{\theta}_u^S$ and $\bm{\theta}_u^D$ trained on unlearnable data can be found in Fig.~\ref{curves}.
    }}
    \label{arch}
\end{center}
\vskip -0.2in
\end{figure}

\begin{figure}[t]
\begin{center}
    \centering
    \includegraphics[width=0.47\textwidth,height=0.35\textwidth]{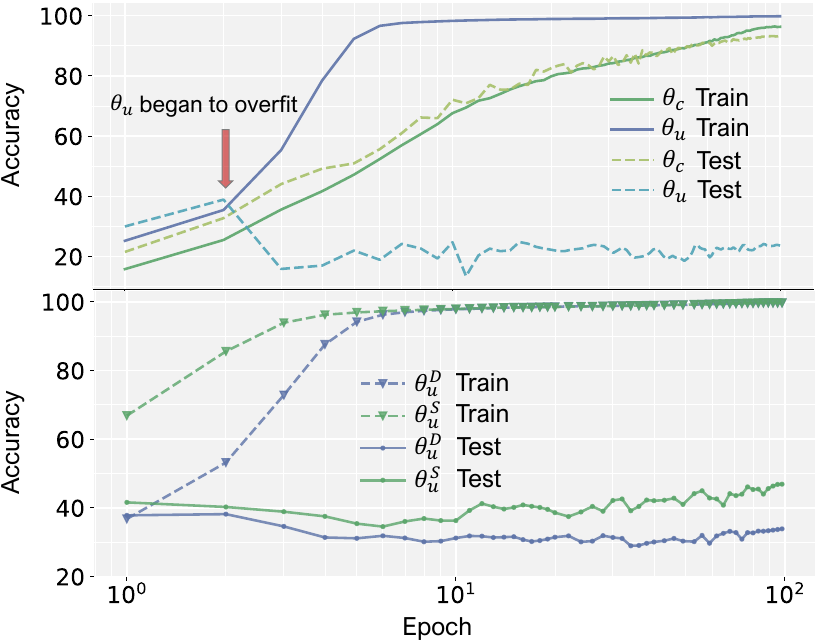}
    \vspace{-0.1in}
    \caption{\footnotesize{\textbf{Top}: the training and test accuracy of a ResNet-18 trained on clean data ($\bm{\theta}_c$) and unlearnable data ($\bm{\theta}_u$). As indicated by the orange arrow, after epoch 3 the training accuracy of $\bm{\theta}_u$ increases sharply, while the test accuracy of $\bm{\theta}_u$ significantly decreases, implying $\bm{\theta}_u$ is trapped in the unlearnable perturbation features. 
    \textbf{Bottom}: the training and test accuracy of $\bm{\theta}_u^S$ (green) and $\bm{\theta}_u^D$ (blue) on unlearnable data. $\bm{\theta}_u^S$ performs better accuracy proving that shallow layers are more crucial for a model to learn correct features during training.}}
    \label{curves}
\end{center}
\vskip -0.25in
\end{figure}

\section{Method}
In this section, we first present our observations and insights into the training process involving unlearnable examples. Our findings reveal two key insights into the learning process of models trained on unlearnable examples. Building upon these insights, we introduce Progressive Staged Training (ST), a novel training framework that learns image semantic features from unlearnable examples.
We further propose  Activation Cluster Measurement (\textit{ACM}), a novel determinant employed to ascertain whether the comprehensive model is succumbing to learning unlearnable perturbation features.

\begin{figure*}[t]
\begin{center}
	\includegraphics[width=\linewidth]{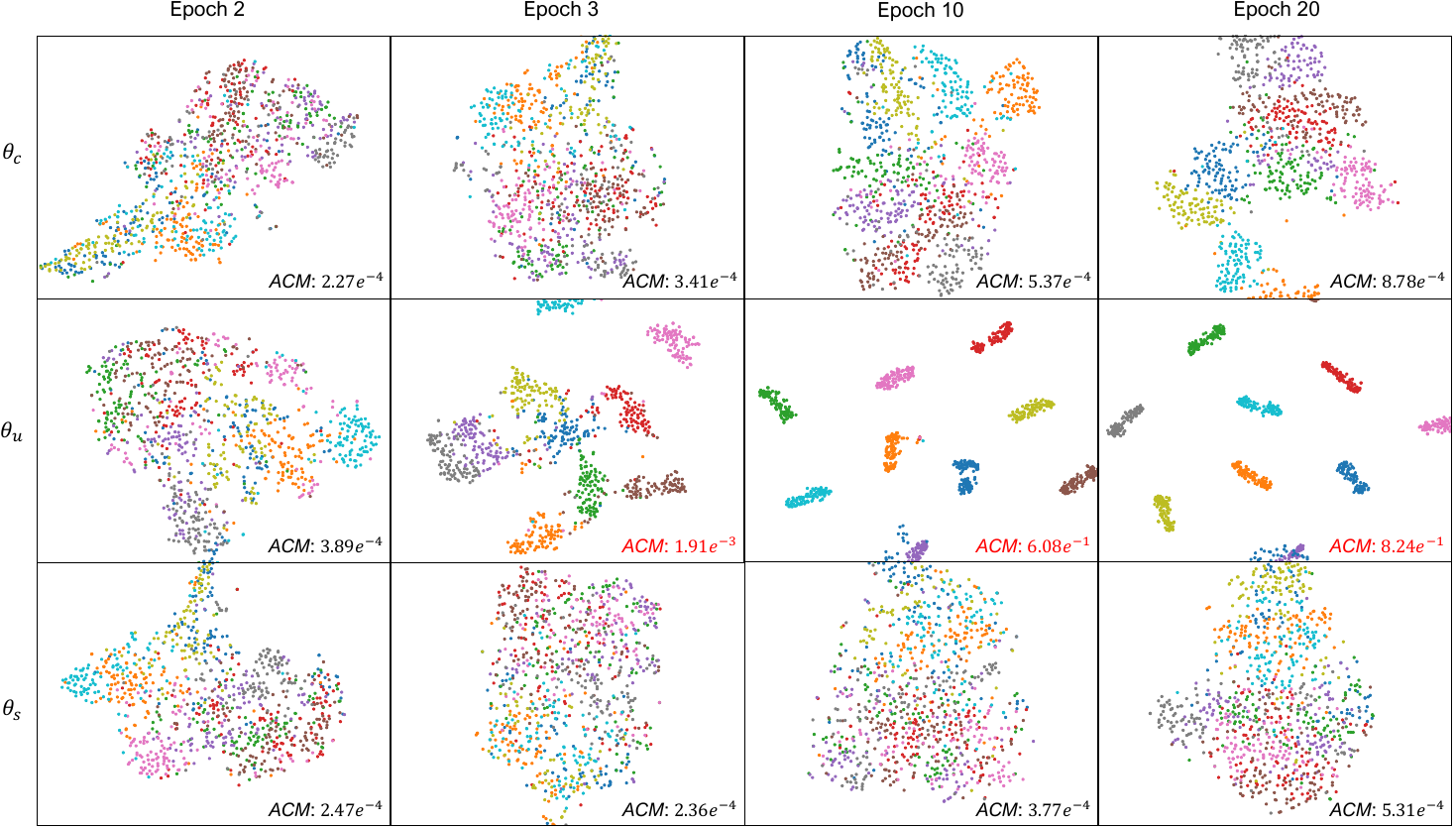}
    \vspace{-0.2in}
\caption{\footnotesize{The penultimate layer activation t-SNE results and \textit{ACM} (~Eq.\eqref{eq:acm}) of various models in different epochs. $\bm{\theta}_c$ is a model naturally trained on clean data. $\bm{\theta}_u$ is a model naturally trained on unlearnable data. $\bm{\theta}_s$ is a model ST trained on unlearnable data. The t-SNE results and \textit{ACM} of $\bm{\theta}_s$ are similar to $\bm{\theta}_c$ and different from $\bm{\theta}_u$, which reflects that our method ST truly prevents models from learning unlearnable perturbation features.}}
\label{t-SNE}
\end{center}
\vskip -0.2in
\end{figure*}

\subsection{Observations and Insights}
\label{insight}

\emph{Insight 1: Model learns both image semantic feature and unlearnable perturbation features at the \textbf{early} stage before being trapped in unlearnable perturbation feature learning, which provides a ``golden window'' for image semantic feature learning.} 
To demonstrate this insight, we show the learning process of a ResNet-18 on unlearnable CIFAR-10 in Fig.~\ref{curves} (more results of other architectures and datasets can be found in Sec. \ref{Observations_for_insights}). 
The generation of unlearnable perturbations is designed to make it easier for the model to achieve its learning objective (a lower loss function value). As a result, the model develops a preference for these perturbation features, which act as ``lures", causing it to neglect the learning of semantic features from the images. This mechanism serves as a means of data protection. As shown in Fig.~\ref{curves}, the training accuracy of the model continues to increase, while the testing accuracy remains at a low level. However, we observe that training accuracy and test accuracy increase during the initial few epochs. This suggests that the model successfully acquires valid semantic features of the image. After the third epoch, the training accuracy increases sharply, while the test accuracy experiences a significant decline. This indicates that the model is becoming trapped in unlearnable perturbation features learning.

This insight motivates us to dynamically adjust the model's learning process during training to reduce the chance of learning unlearnable perturbation features. To achieve this, it is first necessary to identify the point at which the model becomes trapped in learning unlearnable perturbation features. In this work, we propose \textit{(ACM)} to detect this point. We train the model under the monitoring of a metric \textit{(ACM)} to prevent it from being captured by unlearnable perturbation features.

We also consider whether counter-overfitting methods (such as Cutout, Mixup, Cutmix, Drop-out, Weright-decay, Auto-augment, Gaussian-filter)~\cite{dropout,weight-decay,cutout,mixup,cutmix,autoaug,randaug} can effectively liberate the model from learning unlearnable perturbation features (details in Sec.~\ref{c-overfitting}). However, these methods do not account for such powerful unlearnable perturbations and consequently, underperform in overcoming unlearnable examples.

\emph{Insight 2: Compared to deeper layers, correctly guiding the model's \textbf{shallow} layers to learn image semantic features plays a crucial role in preventing the model from becoming trapped in learning unlearnable perturbation features.} 
As shown in Fig.~\ref{arch}, to demonstrate such insight, we first trained a ResNet-18~\cite{resnet} model on clean CIFAR-10~\cite{cifar} denoted by $\bm{\theta}_{c}$, which achieves 92.41\% accuracy on CIFAR-10 test data. 
Then, a ResNet-18 was randomly initialized and the first two residual blocks were supplanted with the initial two residual blocks from $\bm{\theta}_{c}$. The parameters of these replaced blocks were then set to be unmodifiable. This modified model is denoted as $\bm{\theta}_u^S$.
Similarly, we randomly initialized another ResNet-18 and replaced the last two residual blocks with the last two residual blocks of $\bm{\theta}_{c}$. Then we froze these replaced blocks' parameters and denoted this ResNet18 by $\bm{\theta}_u^D$. 
Finally, we trained $\bm{\theta}_u^S$ and $\bm{\theta}_u^D$ on CIFAR-10 unlearnable data individually. Training results are presented in Fig.~\ref{curves} (more results of other architectures and datasets can be found in Sec. \ref{Observations_for_insights}). 

Our findings demonstrate that $\bm{\theta}_u^S$ performs better on CIFAR-10 test data, which proves that we should concentrate on the training process of a model's shallow layers if we aim to overcome unlearnable examples (like $\bm{\theta}_u^S$). This is because incorrect activation signals transmitted from the shallow layers can easily mislead the deeper layers, even if the deep layers themselves correctly learn image semantic features (like $\bm{\theta}_u^D$), which leads the whole model trapped in unlearnable perturbation features learning. 

In this case, correctly guiding the model's shallow layers to learn image semantic features is crucial for undermining the protective capability of unlearnable examples. 
This insight prompted us to implement a learning rate scheduling strategy. We gradually diminish the learning process from the shallow to the deeper layers of the model to retain the image semantic features already learned by the model and break the learning trap created by unlearnable perturbations.

\begin{algorithm}[tb]
   \caption{\footnotesize{Prorgessive Staged Training (ST)}}
   \label{algo:st}
\begin{algorithmic}[1]
   \State {\bfseries Input:} dataset $\mathcal{D}$, model $f_{\bm{\theta}}=\{f_{\bm{\theta}_1},f_{\bm{\theta}_2},\ldots,f_{\bm{\theta}_l}\}$, subset $\mathcal{D}^{s}$, threshold $\gamma$, hyperparameter $\beta$, epoch number $E$, learning rate $\eta$, attenuation function $Atten$.
   \State \textbf{Initialize} $\tau_e\leftarrow0$, $\hat{\eta}^0_i\leftarrow\eta$, $\bm{H}^{tmp}\leftarrow\bm{H}^0\leftarrow[\alpha^0_1,\alpha^0_2,\ldots,\alpha^0_l]\leftarrow[1,1,\ldots,1]$, $\bm{\theta}^{tmp}\leftarrow\bm{\theta}$.
   \For{$e=1$ {\bfseries to} $E$}
        \State $\hat{\eta}^e_i\leftarrow f_{\textit{Atten}}(\hat{\eta}^{e-1}_i)$, $\eta^e_i\leftarrow\alpha^e_i\cdot\hat{\eta}^e_i$ 
        \State $\bm{\theta}_i\leftarrow \textit{Train}(\eta^e_i,\mathcal{D})$   
        \If{$f_{\textit{ACM}}(\bm{\theta},\mathcal{D}^s)>\gamma$ and $\tau^e<{\frac{1}{\beta}-1}$}  
            \State $\tau^e\leftarrow\tau^e+1$ \Comment{tends to learn perturbation features}
            \State $\bm{\theta}\leftarrow\bm{\theta}^{tmp}$   \Comment{load checkpoint from $\bm{\theta}^{tmp}$}
            \State $\bm{H}^e\leftarrow[\alpha^e_1,\alpha^e_2,\ldots,\alpha^e_l]$  
            \Comment{ Eq.~\eqref{eta}}
        \Else  \Comment{model does not tend to learn perturbation features}
            \If{$\tau^e=0$}
                \State $\bm{H}^e\leftarrow[1,1,\ldots,1]$  \Comment{shift to natural training}
            \Else
                \State $\bm{H}^e\leftarrow\bm{H}^{tmp}$  \Comment{adjust learning rate}
            \EndIf
            \State $\bm{\theta}^{tmp}\leftarrow\bm{\theta}$ \Comment{save checkpoint to $\bm{\theta}^{tmp}$}
        \EndIf
        \State $\bm{H}^{tmp}\leftarrow\bm{H}^e$
    \EndFor
\end{algorithmic}
\end{algorithm}

\subsection{ST Training Framework}
\label{sft}
In light of the aforementioned insights, we introduce Progressive Staged Training (ST), a novel strategy that gradually adjusts the learning rate of different layers, moving from shallow to deep layers, when shows a tendency to learn unlearnable perturbation features. To determine the appropriate timing for the modification of the learning rate of these layers, we propose the Activation Cluster Measurement (\textit{ACM}). It serves as an indicator to identify the point at which the model becomes trapped in learning unlearnable perturbation features during the training phase. Simultaneously, we propose a novel learning rate schedule algorithm progressively curbing the shallow layers' unlearnable perturbation features learning process. A comprehensive overview of ST can be found in Fig.~\ref{overview}.

\vspace{1mm}
\textbf{\textit{ACM} Indicator.}
We propose a metric to quantitatively estimate whether the model is in a perturbation learning state, based on the observation that unlearnable perturbations construct simple features, and models are recognized to be “lazy” and prefer to learn simple unlearnable perturbation features rather than difficult image semantic features~\cite{SP}.

To make the above analysis clear and visualized, we present the t-SNE~\cite{tSNE} clustering results of the penultimate layer activation at different training epochs. As shown in Fig.~\ref{t-SNE}, when a model is trained on unlearnable examples, the activation cluster disorder at early epochs, i.e., the intra-class distance is large and inter-class distance is small. As training goes on, the activation begins to cluster well (a small intra-class distance and a large inter-class distance), which is a phenomenon of unlearnable perturbation features learning (other t-SNE results can be found in Sec. \ref{ACM_analysis}). 
To quantify this phenomenon, we propose a clustering measure named Activation Cluster Measurement (\textit{ACM}) to describe the disorder of activation in different training epochs inspired by \cite{TUE}. Suppose that $\mathcal{D}=\{{\mathcal{D}_i}\}_{i=1}^k$ is a dataset with $k$ labels, and ${\mathcal{D}_i}$ is the set of samples in class $i$.
We define set ${\bm{A}_{i}^{\bm{\theta}}}$ as the penultimate layer activation of model $\bm{\theta}$ when given it with set ${\mathcal{D}_i}$ as inputs. 
The cluster center of class $i$ is defined as $\bm{\zeta}\left(i\right)=\frac{1}{\left|{\bm{A}}_i^{\bm{\theta}}\right|}\sum_{{\bm{u}}\in {\bm{A}}_{i}^{\bm{\theta}}}\bm{u}$. Then we have the intra-class distance $f_{\textit{intra}}$ for class $i$ as:
\begin{equation}
    f_{\textit{intra}}\left(i\right)=\frac{1}{\left|{\bm{A}}_i^{\bm{\theta}}\right|}\sum_{{\bm{u}}\in {\bm{A}}_{i}^{\bm{\theta}}}\Vert \bm{u}-\bm{\zeta}(i)\Vert_2
\end{equation}
And the inter-class distance $f_{inter}$ between class $i$ and $j$ is as follows:
\begin{equation}
    f_{\textit{inter}}\left(i,j\right)=\min{\left\{\Vert \bm{u}-\bm{v}\Vert_2\right\}}_{\bm{u}\in \bm{A}_i^{\bm{\theta}},\bm{v}\in \bm{A}_j^{\bm{\theta}}}
\end{equation}
Then we define \textit{ACM} of model $\theta$ on dataset $D$ as:
\begin{equation}
    f_{\textit{ACM}}(\bm{\theta},\mathcal{D})=\frac{1}{k(k-1)}\sum_{\substack{i,j=1\\i\neq j}}^{k}{\frac{f_{\textit{inter}}(i,j)}{\xi_i f_{\textit{intra}}(i)+\xi_jf_{\textit{intra}}(j)}}
\label{eq:acm}
\end{equation}
where $k$ is the number of classes. $\xi_i$ is the radius of class $i$: $\xi_i=\max\left\{\Vert\ \bm{u} - \bm{\zeta}\left(i\right)\Vert_2\right\}_{\bm{u}\in \bm{A}_i^{\bm{\theta}}}$. We use $\mathcal{D}_s$, a subset of the training dataset, as the validation set to calculate \textit{ACM} metric during staged training. 
As the training progresses, the clustering of the model's activation values becomes more pronounced, with the inter-cluster distance ($f_{inter}$) increasing, and the intra-cluster distance ($f_{intra}$)) decreasing, leading to a larger \textit{Activation Clustering Metric (ACM)} value. By observing the \textit{ACM}, we can accurately measure the extent of unlearnable perturbation features learning in the model, which is beneficial for taking steps to learn image semantic features. 

\begin{figure}[t]
\vspace{-0.1in}
\begin{center}
\includegraphics[width=0.47\textwidth]{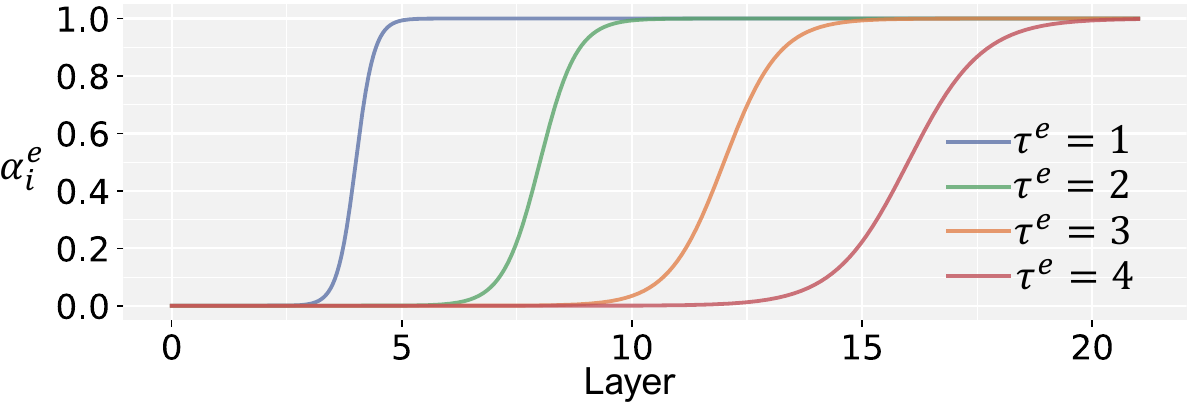}
\end{center}
\vspace{-0.2in}
\caption{\footnotesize{Different $\tau^e$ value for $\alpha^e_i$ in Eq.~\eqref{eta} with $\beta=\frac{1}{5}$, $l=20$. When a model trends to learn unlearnable perturbation features the value of $\tau^e$ will increase and the $\alpha^e_i$ of different layers will decrease, which means the learning rate of these layers will decrease gradually to terminate incorrect feature learning.}}
\label{plot_fun}
\vspace{-0.25in}
\end{figure}

\vspace{1mm}
\textbf{Progressive Staged Training.}~
Based on the insights, we propose progressive staged training (ST), a novel stage training framework to defeat unlearnable examples. For a given model $f_{\bm{\theta}}=\{f_{\bm{\theta}_1},f_{\bm{\theta}_2},\ldots,f_{\bm{\theta}_l}\}$ with $l$ layers and a dataset $\mathcal{D}$, the natural training (NT) process goes through three steps traditionally in each epoch: learning rate attenuation, forward propagation, and backpropagation. Differently, ST goes through five steps in each epoch: learning rate attenuation, learning rate adjustment, forward propagation, backpropagation, and \textit{ACM} calculation. Once \textit{ACM} indicates unlearnable perturbation features learning at the end of an epoch, the model will roll back to the checkpoint of the last epoch. Then, the learning rate will be further modified by an adjustment algorithm after learning rate attenuation to slow down the learning process of shallow layers gradually to resist the wrong learning process. Suppose the initial learning rate of the $i$th layer of model $\bm{\theta}$ is $\hat{\eta}^0_i$ and $f_{\textit{Atten}}$ is a learning rate attenuation function like momentum~\cite{momentum}, cosine~\cite{cosine}, etc. In every epoch, we first get an attenuated learning rate $\hat{\eta}^e_i=f_{\textit{Atten}}(\hat{\eta}^{e-1}_i)$, $e$ for the $e$-th epoch. Then we get a new learning rate $\eta^e_i$ by adjusting $\hat{\eta}^e_i$ with $\eta^e_i=\alpha^e_i\cdot\hat{\eta}^e_i$ ($i$ for the $i$-th layer).
\begin{equation}
    \alpha^e_i=1-\frac{1}{1+\mathrm{e}^{\frac{i}{\beta\cdot \tau^{e}}-l}}
\label{eta}
\end{equation}
where $\tau^e$ is a counter for how many time \textit{ACM} larger than a threshold $\gamma$ from the beginning to epoch $e$ and $\beta$ is a hyperparameter. 
As in Fig.~\ref{plot_fun}, when a model inclines towards learning unlearnable perturbation features, the value of $\tau^e$ escalates concurrently with a reduction in the values of $\alpha^e_i$ across a greater number of shallow layers. This implies that the learning rate associated with these layers will gradually reduce, thereby terminating the learning of incorrect features. 

The process of ST is described in Alg.~\ref{algo:st}. 
We use a vector $\bm{H}^e=[\alpha^e_1,\alpha^e_2,\ldots,\alpha^e_l]$ to describe how the learning rate adjustment algorithm works and defeats unlearnable example. At the beginning of training, we init $\bm{H}^0=[1,1,\ldots,1]$ (line 2), which is equivalent to natural training, and all layers have the same learning rate. As the training goes on, the model tends to learn unlearnable perturbation features detected by \textit{ACM}, and the count $\tau_e$ is increased (line 7). 
Subsequently, the model reverts to a previous checkpoint where it has not yet become trapped in learning unlearnable perturbation features (line 8). As $\tau_e$ increases, the $\alpha^e_i$ of shallow layers decreases gradually, and $\bm{H}^e$ consequently logs $\alpha^e_i$ for each layer of the model (line 9). Following this, the next training epoch commences, with the $\alpha^e_i$ stored in $\bm{H}^e$ adjusting the decayed learning rate (line 4). 
Therefore, following the components of $\mathbf{H}$, the learning rate of each layer gradually decreased to zero, which means the optimization gradually slowed down from shallow layers to deep layers to resist the wrong learning process.

\vspace{1mm}
\textbf{ST Training Pipeline.}
We also investigate that some data augmentations, e.g., color-jitter and gray-scale (CG)~\cite{cg}, could promote the performance of ST (details in Sec. \ref{whycg}). We analyze that CG makes unlearnable perturbations harder to learn for models and weaker on data protection, even if including these augmentations into the EoT process~\cite{REM} of some unlearnable example generators. Furthermore, we analyze adaptive attacks that impose progressive staged training while generating perturbations (details in Sec. \ref{adaptive_attack}). 
The results show that such adaptive attack perturbations are weak in data protection, and progressive staged training works well on them.

In this case, we propose the complete ST training pipeline. This involves two principal steps. In step one, we implement progressive staged training alongside augmentation CG to train a model. In step two, the model developed in step one undergoes fine-tuning aided by augmentation CG. Utilizing this ST training pipeline, we achieve state-of-the-art results to defeat unlearnable examples.

\section{Experiments}
\subsection{Experimental Setting and Implementation}
\subsubsection{Datasets and model architectures} 
Three benchmarks, CIFAR-10, CIFAR-100~\cite{cifar}, and ImageNet-mini, a subset of the first 100 classes in ImageNet~\cite{imagenet}, are used in our experiments, which is consistent with previous works of unlearnable example~\cite{EM, REM, SP, OPS}. We demonstrate the effectiveness of ST with various model architectures including ResNet-18, ResNet-50~\cite{resnet}, WideResNet-28-10 (WRN-28), WideResNet-34-10 (WRN-34)~\cite{wrn}, VGG-16~\cite{vgg16-bn}, and DenseNet-121~\cite{densenet121}.

\begin{table}[t]
\caption{\footnotesize{Notions of different training methods, where NT denotes "natural training" which is our baseline, ST denotes our "staged training".}}
\label{ST-method}
\vspace{-0.15in}
\begin{center}
\adjustbox{width=0.47\textwidth}{
\begin{tabular}{lcccccc}
\toprule
    & \normalfont{NT} & \normalfont{NT-CG} & \normalfont{ST} & \normalfont{ST-CG} & \normalfont{ST-Full} \\
\midrule
Learning rate adjustment & - & - & \checkmark & \checkmark & \checkmark \\
Augmentation CG & $\times$ & \checkmark & $\times$ & \checkmark & \checkmark \\
Fine-tuning with CG & - & $\times$ & $\times$ & $\times$ & \checkmark \\
\bottomrule
\end{tabular}
}
\end{center}
\vspace{-0.25in}
\end{table}

\begin{table*}[t]
    \vspace{-0.1in}
    \caption{\footnotesize{CIFAR-10 clean test accuracies (\%) and average run time ($h$) for natural training (NT), adversarial training (AT), UEraser, LE and ST in different versions (ours) with various models on four unlearnable example methods.}}
    \label{r_CIFAR-10}
    \vspace{-0.2in}
    \begin{center}
    \adjustbox{width=0.9\textwidth}{
    \belowrulesep=0pt
    \aboverulesep=0pt
    \begin{tabular}{l|c|ccccccc|c}
        \toprule
            & Methods  & ResNet-18 & ResNet-50 & VGG-16 & DenseNet-121 & WRN-28 & WRN-34 & Avg. Acc.(\%) & Avg. Time (h)\\
        \midrule
        \multirow{7}{*}{Clean} & \normalfont{NT}  & 92.41 & 91.78 & 91.57 & \textbf{95.04} & 94.68 & \textbf{93.69} & 93.20 & 0.76\\
        & \normalfont{AT} & 88.23 & 87.96 & 85.32 & 44.03 & 78.18 & 84.01 & 77.96 \textcolor{red!60!black}{(-15.24)} & 5.27\\
        & \cellcolor{gray!30} \normalfont{ST} & \cellcolor{gray!30} 82.43 & \cellcolor{gray!30} 84.16 & \cellcolor{gray!30} 70.00 & \cellcolor{gray!30} 59.02 & \cellcolor{gray!30} 91.55 & \cellcolor{gray!30} 84.11 & \cellcolor{gray!30} 78.55 \textcolor{red!60!black}{(-14.66)} & \cellcolor{gray!30} 2.01\\
        & \cellcolor{gray!30} \normalfont{ST-CG} & \cellcolor{gray!30} 86.62 & \cellcolor{gray!30} 86.82 & \cellcolor{gray!30} 80.56 & \cellcolor{gray!30} 62.32 & \cellcolor{gray!30} 92.06 & \cellcolor{gray!30} 85.44 & \cellcolor{gray!30} 82.30 \textcolor{red!60!black}{(-10.90)} & \cellcolor{gray!30} 2.31\\
        & \cellcolor{gray!30} \normalfont{ST-Full} & \cellcolor{gray!30}  \textbf{93.79} & \cellcolor{gray!30} \textbf{94.73} & \cellcolor{gray!30} \textbf{93.11} & \cellcolor{gray!30} 89.47 & \cellcolor{gray!30} \textbf{95.37} & \cellcolor{gray!30} 93.05 & \cellcolor{gray!30} \textbf{93.25 \textcolor{green!60!black}{(+0.05)}} & \cellcolor{gray!30} 3.47\\
        \midrule
        \multirow{7}{*}{EM} & NT  & 19.93 & 18.89 & 14.88 & 20.25 & 16.79 & 38.26 & 21.50 & 0.71\\
        & \normalfont{AT} & 88.62 & 89.28 & \textbf{86.28} & 46.02 & 71.67 & 90.05 & 78.65 \textcolor{green!60!black}{(+57.15)} & 5.30\\
        & \normalfont{LE} & 81.51 & 76.17 & 84.28 & 77.14 & 83.29 & 84.36 & 81.13 \textcolor{green!60!black}{(+59.63)} & 2.22\\
        & \normalfont{UEraser} & 86.11 & 85.59 & 81.76 & 86.12 & 81.56 & 86.44 & 84.60\textcolor{green!60!black}{(+63.10)} & 1.51\\
        & \cellcolor{gray!30} \normalfont{ST} & \cellcolor{gray!30} 58.63 & \cellcolor{gray!30} 78.44 & \cellcolor{gray!30} 69.97 & \cellcolor{gray!30} 56.65 & \cellcolor{gray!30} 37.43 & \cellcolor{gray!30} 79.43 & \cellcolor{gray!30} 63.425 \textcolor{green!60!black}{(+41.93)} & \cellcolor{gray!30} 1.98\\
        & \cellcolor{gray!30} \normalfont{ST-CG} & \cellcolor{gray!30} 80.18 & \cellcolor{gray!30} 83.43 & \cellcolor{gray!30} 61.52 & \cellcolor{gray!30} 58.11 & \cellcolor{gray!30} 84.56 & \cellcolor{gray!30} 86.56 & \cellcolor{gray!30} 75.73 \textcolor{green!60!black}{(+54.23)} & \cellcolor{gray!30} 2.34\\
        & \cellcolor{gray!30} \normalfont{ST-Full} & \cellcolor{gray!30} \textbf{87.05} & \cellcolor{gray!30} \textbf{92.84} & \cellcolor{gray!30} 82.30 & \cellcolor{gray!30} \textbf{87.18} & \cellcolor{gray!30} \textbf{88.69} & \cellcolor{gray!30} \textbf{93.11} & \cellcolor{gray!30} \textbf{88.53 \textcolor{green!60!black}{(+67.03)}} & \cellcolor{gray!30} 3.61\\
        \midrule
        \multirow{7}{*}{REM} & NT & 21.86 & 25.30 & 38.81 & 32.28 & 26.43 & 20.37 & 27.51 & 0.76\\
        & \normalfont{AT} & 48.16 & 40.65 & 65.23 & 32.04 & 26.92 & 48.39 & 43.57\textcolor{green!60!black}{(+16.06)} & 5.05\\
        & \normalfont{LE} & 80.78 & 80.67 & 72.91 & 70.29 & 75.22 & 78.27 & 76.36 \textcolor{green!60!black}{(+48.85)} & 2.65\\
        & \normalfont{UEraser} & 67.48 & 78.18 & 71.59 & 71.91 & 61.75 & 61.43 & 68.78 \textcolor{green!60!black}{(+41.27)} & 1.62\\
        & \cellcolor{gray!30} \normalfont{ST} & \cellcolor{gray!30} \textbf{82.11} & \cellcolor{gray!30} 68.90 & \cellcolor{gray!30} 57.13 & \cellcolor{gray!30} 41.23 & \cellcolor{gray!30} 39.84 & \cellcolor{gray!30} 81.25 & \cellcolor{gray!30} 61.74 \textcolor{green!60!black}{(+34.23)} & \cellcolor{gray!30} 2.00\\
        & \cellcolor{gray!30} \normalfont{ST-CG} & \cellcolor{gray!30} 77.08 & \cellcolor{gray!30} 79.02 & \cellcolor{gray!30} 61.03 & \cellcolor{gray!30} 43.56 & \cellcolor{gray!30} \textbf{76.16} & \cellcolor{gray!30} 83.04 & \cellcolor{gray!30} 69.98 \textcolor{green!60!black}{(+42.47)} & \cellcolor{gray!30} 2.01\\
        & \cellcolor{gray!30} \normalfont{ST-Full} & \cellcolor{gray!30} 78.86 & \cellcolor{gray!30} \textbf{85.86} & \cellcolor{gray!30} \textbf{74.43} & \cellcolor{gray!30} \textbf{72.94} & \cellcolor{gray!30} 74.23 & \cellcolor{gray!30} \textbf{84.01} & \cellcolor{gray!30} \textbf{78.39 \textcolor{green!60!black}{(+50.88)}} & \cellcolor{gray!30} 3.17\\
        \midrule
        \multirow{7}{*}{SP} & NT & 13.67 & 15.85 & 16.40 & 25.83 & 14.31 & 12.95 & 16.50 & 0.78\\
        & \normalfont{AT} & 22.21 & 52.02 & 61.12 & 47.63 & 58.26 & 64.41 & 50.94 \textcolor{green!60!black}{(+34.44)} & 5.27\\
        & \normalfont{LE} & 82.94 & \textbf{86.78} & 83.78 & \textbf{76.73} & 83.27 & 81.48 & 82.50 \textcolor{green!60!black}{(+66.00)} & 2.19\\
        & \normalfont{UEraser} & \textbf{80.58} & 85.57 & 84.56 & 75.45 & 85.65 & 83.47 & 82.55 \textcolor{green!60!black}{(+66.05)} & 1.62\\
        & \cellcolor{gray!30} \normalfont{ST} & \cellcolor{gray!30} 42.73 & \cellcolor{gray!30} 39.43 & \cellcolor{gray!30} 53.31 & \cellcolor{gray!30} 50.59 & \cellcolor{gray!30} 37.25 & \cellcolor{gray!30} 42.04 & \cellcolor{gray!30} 44.23 \textcolor{green!60!black}{(+27.73)} & \cellcolor{gray!30} 1.84\\
        & \cellcolor{gray!30} \normalfont{ST-CG} & \cellcolor{gray!30} 66.75 & \cellcolor{gray!30} 77.19 & \cellcolor{gray!30} 79.43 & \cellcolor{gray!30} 55.36 & \cellcolor{gray!30} 86.32 & \cellcolor{gray!30} 79.82 & \cellcolor{gray!30} 74.15 \textcolor{green!60!black}{(+57.65)} & \cellcolor{gray!30} 1.86\\
        & \cellcolor{gray!30} \normalfont{ST-Full} & \cellcolor{gray!30} 79.69 & \cellcolor{gray!30} 83.75 & \cellcolor{gray!30} \textbf{86.13} & \cellcolor{gray!30} 73.15 & \cellcolor{gray!30} \textbf{87.25} & \cellcolor{gray!30} \textbf{85.47} & \cellcolor{gray!30} \textbf{82.57 \textcolor{green!60!black}{(+66.07)}} & \cellcolor{gray!30} 2.98\\
        \midrule
        \multirow{7}{*}{OPS} & NT & 21.71 & 19.20 & 17.25 & 25.01 & 31.67 & 19.69 & 22.42 & 0.82\\
        & \normalfont{AT} & 21.08 & 15.20 & 18.6 & 32.14 & 48.19 & 47.75 & 30.49 \textcolor{green!60!black}{(+8.07)} & 5.06\\
        & \normalfont{LE} & 75.59 & 71.91 & 63.16 & 65.47 & \textbf{71.22} & 76.91 & 70.71 \textcolor{green!60!black}{(+48.29)} & 2.51\\
        & \normalfont{UEraser} & 69.75 & 68.40 & 56.95 & \textbf{68.15} & 61.68 & 68.76 & 65.62 \textcolor{green!60!black}{(+43.20)} & 1.68\\
        & \cellcolor{gray!30} \normalfont{ST} & \cellcolor{gray!30} 53.25 & \cellcolor{gray!30} 57.18 & \cellcolor{gray!30} 44.66 & \cellcolor{gray!30} 55.02 & \cellcolor{gray!30} 41.45 & \cellcolor{gray!30} 73.55 & \cellcolor{gray!30} 54.19 \textcolor{green!60!black}{(+31.77)} & \cellcolor{gray!30} 1.87\\
        & \cellcolor{gray!30} \normalfont{ST-CG} & \cellcolor{gray!30} 75.58 & \cellcolor{gray!30} 80.37 & \cellcolor{gray!30} \textbf{65.45} & \cellcolor{gray!30} 58.06 & \cellcolor{gray!30} 64.84 & \cellcolor{gray!30} 81.34 & \cellcolor{gray!30} \textbf{70.94 \textcolor{green!60!black}{(+48.52)}} & \cellcolor{gray!30} 1.92\\
        & \cellcolor{gray!30} \normalfont{ST-Full} & \cellcolor{gray!30} \textbf{76.19} & \cellcolor{gray!30} \textbf{80.93} & \cellcolor{gray!30} 57.16 & \cellcolor{gray!30} 45.13 & \cellcolor{gray!30} 64.79 & \cellcolor{gray!30} \textbf{81.86} & \cellcolor{gray!30} 67.68 \textcolor{green!60!black}{(+45.26)} & \cellcolor{gray!30} 3.06\\
        \midrule
        \multirow{7}{*}{AEP} & NT & 6.82 & 9.17 & 13.08 & 22.43 & 13.91 & 9.78 & 12.53 & 0.74 \\
        & \normalfont{AT} & 73.84 & 71.26 & 58.68 & 38.88 & \textbf{74.81} & 76.95 & 65.74 \textcolor{green!60!black}{(+53.21)} & 5.31\\
        & \normalfont{LE} & 73.22 & 74.62 & 60.27 & 48.65 & 59.83 & 62.76 & 63.23 \textcolor{green!60!black}{(+50.70)} & 2.47\\
        & \normalfont{UEraser} & 51.13 & 44.97 & 39.37 & 52.52 & 66.83 & 49.72 & 50.76 \textcolor{green!60!black}{(+38.23)} & 1.49\\
        & \cellcolor{gray!30} \normalfont{ST} & \cellcolor{gray!30} 82.48 & \cellcolor{gray!30} 72.81 & \cellcolor{gray!30} \textbf{64.56} & \cellcolor{gray!30} \textbf{55.53} & \cellcolor{gray!30} 45.71 & \cellcolor{gray!30} 72.29 & \cellcolor{gray!30} 65.56 \textcolor{green!60!black}{(+53.03)} & \cellcolor{gray!30} 1.78\\
        & \cellcolor{gray!30} \normalfont{ST-CG} & \cellcolor{gray!30} 84.45 & \cellcolor{gray!30} 82.94 & \cellcolor{gray!30} 62.64 & \cellcolor{gray!30} 55.27 & \cellcolor{gray!30} 62.28 & \cellcolor{gray!30} 84.25 & \cellcolor{gray!30} \textbf{71.97 \textcolor{green!60!black}{(+59.44)}} & \cellcolor{gray!30} 1.81\\
        & \cellcolor{gray!30} \normalfont{ST-Full} & \cellcolor{gray!30} \textbf{85.08} & \cellcolor{gray!30} \textbf{83.31} & \cellcolor{gray!30} 51.38 & \cellcolor{gray!30} 46.63 & \cellcolor{gray!30} 60.15 & \cellcolor{gray!30} \textbf{84.72} & \cellcolor{gray!30} 68.55 \textcolor{green!60!black}{(+56.01)} & \cellcolor{gray!30} 3.11\\
        \midrule
        \multirow{7}{*}{AP} & NT & 18.43 & 17.21 & 16.86 & 33.06 & 12.78 & 10.44 & 18.13 & 0.72 \\
        & \normalfont{AT} & 60.13 & 57.21 & 11.21 & 18.67 & 73.65 & 74.58 & 49.24 \textcolor{green!60!black}{(+31.11)} & 5.21 \\
        & \normalfont{LE} & 78.61 & 81.03 & 75.24 & 71.83 & 75.01 & 76.25 & 76.33 \textcolor{green!60!black}{(+58.20)} & 2.52\\
        & \normalfont{UEraser} & 74.22 & 71.51 & 75.58 & \textbf{75.13} & 76.38 & 72.63 & 74.24 \textcolor{green!60!black}{(+56.11)} & 1.60\\
        & \cellcolor{gray!30} \normalfont{ST} & \cellcolor{gray!30} 82.11 & \cellcolor{gray!30} 81.51 & \cellcolor{gray!30} 69.62 & \cellcolor{gray!30} 57.81 & \cellcolor{gray!30} 59.46 & \cellcolor{gray!30} \textbf{87.08} & \cellcolor{gray!30} 72.93 \textcolor{green!60!black}{(+54.80)} & \cellcolor{gray!30} 1.88 \\
        & \cellcolor{gray!30} \normalfont{ST-CG} & \cellcolor{gray!30} 86.77 & \cellcolor{gray!30} 85.79 & \cellcolor{gray!30} 72.15 & \cellcolor{gray!30} 57.59 & \cellcolor{gray!30} 87.18 & \cellcolor{gray!30} 85.63 & \cellcolor{gray!30} 79.19 \textcolor{green!60!black}{(+61.06)} & \cellcolor{gray!30} 1.91 \\
        & \cellcolor{gray!30} \normalfont{ST-Full} & \cellcolor{gray!30} \textbf{87.51} & \cellcolor{gray!30} \textbf{86.62} & \cellcolor{gray!30} \textbf{77.91} & \cellcolor{gray!30} 65.44 & \cellcolor{gray!30} \textbf{88.21} & \cellcolor{gray!30} 86.79 & \cellcolor{gray!30} \textbf{82.08 \textcolor{green!60!black}{(+63.95)}} & \cellcolor{gray!30} 3.31 \\
        \midrule
        \multirow{7}{*}{HYPO} & NT & 12.34 & 13.15 & 16.64 & 19.23 & 10.11 & 10.39 & 13.64 & 0.73\\
        & \normalfont{AT} & 17.11 & 15.61 & 10.12 & 14.93 & 13.29 & 15.93 & 14.50 \textcolor{green!60!black}{(+0.86)} & 5.30\\
        & \normalfont{LE} & 67.57 & 70.88 & 60.81 & 66.72 & 68.57 & 69.17 & 67.29 \textcolor{green!60!black}{(+53.65)} & 2.22\\
        & \normalfont{UEraser} & 58.94 & 58.71 & \textbf{65.84} & \textbf{64.69} & 68.05 & 51.19 & 61.24 \textcolor{green!60!black}{(+47.60)} & 1.68\\
        & \cellcolor{gray!30} \normalfont{ST} & \cellcolor{gray!30} \textbf{81.75} & \cellcolor{gray!30} 59.28 & \cellcolor{gray!30} 50.13 & \cellcolor{gray!30} 49.30 & \cellcolor{gray!30} 38.51 & \cellcolor{gray!30} 82.11 & \cellcolor{gray!30} 60.18 \textcolor{green!60!black}{(+46.54)} & \cellcolor{gray!30} 1.94\\
        & \cellcolor{gray!30} \normalfont{ST-CG} & \cellcolor{gray!30} 73.09 & \cellcolor{gray!30} 78.12 & \cellcolor{gray!30} 50.30 & \cellcolor{gray!30} 48.73 & \cellcolor{gray!30} 66.84 & \cellcolor{gray!30} 85.55 & \cellcolor{gray!30} 67.11 \textcolor{green!60!black}{(+53.47)} & \cellcolor{gray!30} 2.04\\
        & \cellcolor{gray!30} \normalfont{ST-Full} & \cellcolor{gray!30} 73.19 & \cellcolor{gray!30} \textbf{79.36} & \cellcolor{gray!30} 55.46 & \cellcolor{gray!30} 41.46 & \cellcolor{gray!30} \textbf{68.88} & \cellcolor{gray!30} \textbf{86.05} & \cellcolor{gray!30} \textbf{67.40 \textcolor{green!60!black}{(+53.76)}} & \cellcolor{gray!30} 3.31\\
    \bottomrule
    \end{tabular}
    }
    \end{center}
    \vskip -0.25in
\end{table*}

\begin{table*}[tb]
    \vspace{-0.1in}
        \caption{\footnotesize{CIFAR-100 clean test accuracies (\%) for natural training (NT), adversarial training (AT), UEraser, LE and ST in different versions (ours) with various models on four unlearnable example methods.}} 
        \label{r_CIFAR-100}
        \vspace{-0.2in}
        \begin{center}
        \adjustbox{width=0.95\textwidth}{
        \begin{tabular}{lcccccccc}
            \toprule
            Methods    & Frame  & ResNet-18 & ResNet-50 & VGG-16 & DenseNet-121 & WRN-28 & WRN-34 & Average \\
            \midrule
            \multirow{7}{*}{Clean} & \normalfont{NT} & 70.77 & \textbf{73.69} & 67.27 & \textbf{61.99} & \textbf{77.31} & 71.28 & 70.38\\
            & \normalfont{AT} & 61.50 & 65.81 & 56.59 & 28.23 & 48.72 & 64.01 & 54.14\textcolor{red!60!black}{(-16.24)}\\
            & \cellcolor{gray!30} \normalfont{ST} & \cellcolor{gray!30} 49.57 & \cellcolor{gray!30} 55.70 & \cellcolor{gray!30} 41.50 & \cellcolor{gray!30} 30.18 & \cellcolor{gray!30} 66.86 & \cellcolor{gray!30} 61.37 & \cellcolor{gray!30} 50.86 \textcolor{red!60!black}{(-19.52)}\\
            & \cellcolor{gray!30} \normalfont{ST-CG} & \cellcolor{gray!30} 52.31 & \cellcolor{gray!30} 58.68 & \cellcolor{gray!30} 39.22 & \cellcolor{gray!30} 36.22 & \cellcolor{gray!30} 65.23 & \cellcolor{gray!30} 59.84 & \cellcolor{gray!30} 51.92 \textcolor{red!60!black}{(-18.46)}\\
            & \cellcolor{gray!30} \normalfont{ST-Full} & \cellcolor{gray!30} \textbf{71.84} & \cellcolor{gray!30} 73.11 & \cellcolor{gray!30} \textbf{70.83} & \cellcolor{gray!30} 61.76 & \cellcolor{gray!30} 75.08 & \cellcolor{gray!30} \textbf{72.11} & \cellcolor{gray!30} \textbf{70.78 \textcolor{green!60!black}{(+0.41)}}\\
            \midrule
            \multirow{7}{*}{EM} & NT  & 14.81 & 12.19 & 22.18 & 13.71 & 14.48 & 28.20 & 17.60 \\
            & \normalfont{AT} & 64.17 & 66.43 & 56.96 & 26.51 & 46.15 & 68.27 & 54.75 \textcolor{green!60!black}{(+37.15)}\\
            & \normalfont{LE} & 53.72 & 51.89 & 52.59 & 53.36 & 61.23 & 55.56 & 54.73 \textcolor{green!60!black}{(+37.13)}\\
            & \normalfont{UEraser} & 57.27 & 56.81 & 56.91 & 56.43 & 50.61 & 56.25 & 55.71 \textcolor{green!60!black}{(+38.11)}\\
            & \cellcolor{gray!30} \normalfont{ST} & \cellcolor{gray!30} 47.55 & \cellcolor{gray!30} 57.56 & \cellcolor{gray!30} 38.13 & \cellcolor{gray!30} 32.13 & \cellcolor{gray!30} 51.27 & \cellcolor{gray!30} 59.34 & \cellcolor{gray!30} 47.66 \textcolor{green!60!black}{(+30.06)}\\
            & \cellcolor{gray!30} \normalfont{ST-CG} & \cellcolor{gray!30} 51.93 & \cellcolor{gray!30} 54.91 & \cellcolor{gray!30} 40.59 & \cellcolor{gray!30} 41.05 & \cellcolor{gray!30} \textbf{65.01} & \cellcolor{gray!30} 61.30 & \cellcolor{gray!30} 52.47 \textcolor{green!60!black}{(+34.87)}\\
            & \cellcolor{gray!30} \normalfont{ST-Full} & \cellcolor{gray!30} \textbf{64.81} & \cellcolor{gray!30} \textbf{71.94} & \cellcolor{gray!30} \textbf{70.66} & \cellcolor{gray!30} \textbf{61.77} & \cellcolor{gray!30} 41.36 & \cellcolor{gray!30}  \textbf{68.68} & \cellcolor{gray!30} \textbf{63.20 \textcolor{green!60!black}{(+45.60)}}\\
            \midrule
            \multirow{7}{*}{REM} & NT  & 8.71 & 17.28 & 13.51 & 16.04 & 8.83 & 10.65 & 12.50\\
            & \normalfont{AT} & 27.10 & 26.03 & 48.85 & 21.18 & 39.25 & 25.04 & 31.24 \textcolor{green!60!black}{(+18.74)}\\
            & \normalfont{LE} & 50.21 & 53.86 & \textbf{50.91} & \textbf{50.38} & 52.88 & 50.29 & 51.42 \textcolor{green!60!black}{(+38.92)}\\
            & \normalfont{UEraser} & 50.84 & 55.04 & 49.13 & 48.22 & 54.96 & 52.05 & 51.71 \textcolor{green!60!black}{(+39.21)}\\
            & \cellcolor{gray!30} \normalfont{ST} & \cellcolor{gray!30} 47.61 & \cellcolor{gray!30} 35.52 & \cellcolor{gray!30} 29.81 & \cellcolor{gray!30} 24.33 & \cellcolor{gray!30} 27.54 & \cellcolor{gray!30} 36.29 & \cellcolor{gray!30} 33.52 \textcolor{green!60!black}{(+21.02)}\\
            & \cellcolor{gray!30} \normalfont{ST-CG} & \cellcolor{gray!30} 52.94 & \cellcolor{gray!30} 49.45 & \cellcolor{gray!30} 23.92 & \cellcolor{gray!30} 25.35 & \cellcolor{gray!30} 61.11 & \cellcolor{gray!30} 57.02 & \cellcolor{gray!30} 44.97 \textcolor{green!60!black}{(+32.47)}\\
            & \cellcolor{gray!30} \normalfont{ST-Full} & \cellcolor{gray!30} \textbf{54.19} & \cellcolor{gray!30} \textbf{58.53} & \cellcolor{gray!30} 40.15 & \cellcolor{gray!30} 36.36 & \cellcolor{gray!30} \textbf{61.79} & \cellcolor{gray!30} \textbf{59.41} & \cellcolor{gray!30} \textbf{51.73 \textcolor{green!60!black}{(+39.23)}}\\
            \midrule
            \multirow{7}{*}{SP} & NT & 9.54 & 8.51 & 10.57 & 9.45 & 11.13 & 10.36 & 9.93 \\
            & \normalfont{AT} & 24.81 & 29.84 & 28.71 & 23.19 & 39.11 & 43.09 & 31.46 \textcolor{green!60!black}{(+21.53)}\\
            & \normalfont{LE} & \textbf{53.19} & \textbf{55.27} & 61.74 & \textbf{51.89} & 56.29 & 54.71 & \textbf{55.52 \textcolor{green!60!black}{(+45.59)}}\\
            & \normalfont{UEraser} & 51.18 & 55.11 & 56.16 & 50.03 & 56.73 & 53.61 & 53.80 \textcolor{green!60!black}{(+43.87)}\\
            & \cellcolor{gray!30} \normalfont{ST} & \cellcolor{gray!30} 32.54 & \cellcolor{gray!30} 27.12 & \cellcolor{gray!30} 28.84 & \cellcolor{gray!30} 28.73 & \cellcolor{gray!30} 27.82 & \cellcolor{gray!30} 26.35 & \cellcolor{gray!30} 28.57 \textcolor{green!60!black}{(+18.64)}\\
            & \cellcolor{gray!30} \normalfont{ST-CG} & \cellcolor{gray!30} 48.03 & \cellcolor{gray!30} 48.98 & \cellcolor{gray!30} 37.66 & \cellcolor{gray!30} 34.75 & \cellcolor{gray!30} 54.90 & \cellcolor{gray!30} 46.70 & \cellcolor{gray!30} 45.17 \textcolor{green!60!black}{(+35.24)}\\
            & \cellcolor{gray!30} \normalfont{ST-Full} & \cellcolor{gray!30} 52.46 & \cellcolor{gray!30} 52.52 & \cellcolor{gray!30} \textbf{66.13} & \cellcolor{gray!30} 41.66 & \cellcolor{gray!30} \textbf{59.43} & \cellcolor{gray!30} \textbf{55.09} & \cellcolor{gray!30} 54.55 \textcolor{green!60!black}{(+44.62)}\\
            \midrule
            \multirow{7}{*}{OPS} & NT & 13.93 & 12.21 & 8.90 & 12.85 & 14.55 & 11.34 & 12.30\\
            & \normalfont{AT} & 10.85 & 17.07 & 10.17 & 11.15 & 21.86 & 24.82 & 15.99 \textcolor{green!60!black}{(+3.69)}\\
            & \normalfont{LE} & 44.22 & 40.71 & 33.01 & \textbf{36.38} & 43.36 & 41.93 & 39.94 \textcolor{green!60!black}{(+27.64)}\\
            & \normalfont{UEraser} & 32.77 & 31.56 & 32.63 & 31.67 & 32.29 & 32.45 & 32.23 \textcolor{green!60!black}{(+19.93)}\\
            & \cellcolor{gray!30} \normalfont{ST} & \cellcolor{gray!30} 29.98 & \cellcolor{gray!30} 51.79 & \cellcolor{gray!30} \textbf{36.55} & \cellcolor{gray!30} 29.85 & \cellcolor{gray!30} 32.76 & \cellcolor{gray!30} 56.84 & \cellcolor{gray!30} 39.63 \textcolor{green!60!black}{(+27.32)}\\
            & \cellcolor{gray!30} \normalfont{ST-CG} & \cellcolor{gray!30} \textbf{47.73} & \cellcolor{gray!30} 56.53 & \cellcolor{gray!30} 34.81 & \cellcolor{gray!30} 33.75 & \cellcolor{gray!30} 50.34 & \cellcolor{gray!30} 57.44 & \cellcolor{gray!30} \textbf{46.77 \textcolor{green!60!black}{(+34.47)}}\\
            & \cellcolor{gray!30} \normalfont{ST-Full} & \cellcolor{gray!30} 46.73 & \cellcolor{gray!30} \textbf{56.87} & \cellcolor{gray!30} 27.17 & \cellcolor{gray!30} 30.96 & \cellcolor{gray!30} \textbf{50.55} & \cellcolor{gray!30} \textbf{57.69} & \cellcolor{gray!30} 45.00 \textcolor{green!60!black}{(+32.70)}\\
        \bottomrule
        \end{tabular}
        }
        \end{center}
        \vskip -0.25in
    \end{table*}

\subsubsection{Unlearnable examples} Four state-of-the-art unlearnable examples generation methods are used:  Error-Minimizing perturbation (EM)~\cite{EM}, Robust-Error-Minimizing perturbation (REM)~\cite{REM}, Synthetic Perturbation (SP)~\cite{SP}, One-Pixel Shortcut (OPS)~\cite{OPS}. Meanwhile, three data poison attack generation methods are used: Adversarial Examples as Poisons (AEP)~\cite{2-2-1}, Hypocritical perturbations (HYPO)~\cite{HYPO}, Autoregressive Perturbations (AP)~\cite{AR}. We set the $\ell_{\infty}$ bound as $\epsilon=\frac{8}{255}$ for EM, REM, SP, AEP and $\ell_{2}$ bound as $\epsilon=1$ for AP, HYPO to follow their default setting. We set other hyperparameters to the default value in the open-sourced codes of EM~\footnote{\url{https://github.com/HanxunH/Unlearnable-Examples}}, REM~\footnote{\url{https://github.com/fshp971/robust-unlearnable-examples}}, SP~\footnote{\url{https://github.com/dayu11/Availability-Attacks-Create-Shortcuts}}, OPS~\footnote{\url{https://openreview.net/forum?id=p7G8t5FVn2h}}, 
AEP~\footnote{\url{https://github.com/lhfowl/adversarial_poisons}}, 
AP~\footnote{\url{https://github.com/psandovalsegura/autoregressive-poisoning}}, HYPO~\footnote{\url{https://github.com/TLMichael/Delusive-Adversary}}.

\subsubsection{Baselines}
We adopt adversarial training (AT) with a bound of $\ell_{\infty}\leq\frac{4}{255}$ as the baseline following the REM's setting. Meanwhile, we use Ueraser~\footnote{\url{https://github.com/lafeat/ueraser}} and LE~\footnote{\url{https://github.com/jiangw-0/LE_JCDP}} as baselines. These works primarily focus on data processing methods for learning from unlearnable samples. In contrast, our approach is orthogonal, optimizing the learning process through dynamic adjustment of the learning rate to effectively handle such samples. We set hyperparameters to the default value in their open-sourced codes.

\subsubsection{Training settings}
To show our effectiveness, we compare the natural training method with 3 different settings of ST (as summarized in Tab.~\ref{ST-method}). Specifically, we implement ST with $\beta$ set to $50\%$ for CIFAR-10, ImageNet-mini, and $33.33\%$ for CIFAR-100. Meanwhile, we set $\gamma$ to $2.6\times 10^{-4}$. We use a subset of the training dataset (unlearnable perturbation contaminated images), $\mathcal{D}^s$, containing 1000 samples for CIFAR-10 (each class contains 100 samples) and 5000 samples for CIFAR-100, ImageNet-mini (each class contains 50 samples) as the validation set (all the images in validation set are contaminated by unlearnable perturbation) to calculate the \textit{ACM} metric during staged training. 
Following previous work on unlearnable examples (EM, REM, and SP), we perform random flipping on the image in all experiments. Then, we randomly crop the image to 32$\times$32 size for CIFAR-10 and CIFAR-100 and 224$\times$224 size for lmageNet-mini.

\begin{table}[th]
    \vspace{-0.05in}
    \caption{\footnotesize{ImageNet-mini clean test accuracies (\%) for natural training (NT), adversarial training (AT), and our ST in different settings over various unlearnable example methods on ResNet-18 and DenseNet-121.}}
    \label{r-imagnet-mini}
    \belowrulesep=0pt
    \aboverulesep=0pt
    \vspace{-0.18in}
    \begin{center}
    \adjustbox{width=0.48\textwidth}{
    \begin{tabular}{l|c|cccc}
    \toprule
    \normalfont{Methods} & \normalfont{Arch.} & \normalfont{Clean} & \normalfont{EM} & \normalfont{REM} & \normalfont{SP} \\
    \midrule
    \multirow{2}{*}{NT} & \small{ResNet} & 80.66 & 14.81 & 13.74 & 11.10 \\
    & \small{DenseNet} & 80.27 & 6.58 & 18.51 & 12.58 \\
    \midrule
    \multirow{2}{*}{AT} & \small{ResNet} & 65.73 & 50.05 & 50.43 & 52.05 \\
    & \small{DenseNet} & 54.25 & 37.15 & 38.63 & 38.29 \\
    \midrule
    \rowcolor{gray!30} & \small{ResNet} & 77.36 & 43.27 & 47.5 & 75.59 \\
    \rowcolor{gray!30} \multirow{-2}{*}{ST} & \small{DenseNet} & 65.79 & 65.26 & 48.97 & 65.47 \\
    \midrule
    \rowcolor{gray!30} & \small{ResNet} & 78.93 & \textbf{45.81} & 71.29 & 76.47 \\
    \rowcolor{gray!30} \multirow{-2}{*}{ST-CG} & \small{DenseNet} & 69.86 & 64.4 & 59.69 & 68.37 \\
    \midrule
    \rowcolor{gray!30} &  & \textbf{82.50} & 40.54 & \textbf{72.18} & \textbf{79.36} \\ 
    \rowcolor{gray!30} & \multirow{-2}{*}{\small{ResNet}} & \textcolor{green!60!black}{\textbf{+1.84}} & \textcolor{green!60!black}{+25.73} & \textcolor{green!60!black}{+58.44} & \textcolor{green!60!black}{\textbf{+68.26}} \\
    \rowcolor{gray!30} & & \textbf{80.27} & \textbf{68.19} & \textbf{68.95} & \textbf{79.81} \\
    \rowcolor{gray!30} \multirow{-4}{*}{ST-Full} & \multirow{-2}{*}{\small{DenseNet}} & \textcolor{green!60!black}{\textbf{+1.02}} & \textcolor{green!60!black}{\textbf{+61.61}} & \textcolor{green!60!black}{\textbf{+50.44}} & \textcolor{green!60!black}{\textbf{+67.23}} \\
    \bottomrule
    \end{tabular}
    }
    \end{center}
    \vspace{-0.32in}
\end{table}

For normal training (NT), we use the SGD optimizer accompanied by the cosine learning rate decay, with momentum set to 0.9, weight decay set to $10^{-4}$, and learning rate set to 0.1.
For adversarial training (AT), we use the SGD optimizer accompanied by the cosine learning rate decay, with momentum set to 0.9, weight decay set to $10^{-4}$, and learning rate set to 0.1.
For the ST series (ours), we use the SGD optimizer accompanied by the cosine learning rate decay, with momentum set to 0.9, weight decay set to $10^{-4}$, and learning rate set to 0.1 for ST and 0.2 for ST-CG. In the \textit{ACM} calculation stage, we reduced the activation of the penultimate layer to two dimensions with t-SNE. Meanwhile, we use the first ten classes to calculate \textit{ACM} for CIFAR-10 and ImageNet-mini.  Following traditional data augmentation methods in deep learning, we perform color-jitter on images with 80\% probability and gray-scale with 20\% probability. 
For fine-tuning with CG, we fine-tune all layers of a model and use the SGD optimizer accompanied by the cosine learning rate decay, with momentum set to 0.9, and weight decay set to $10^{-4}$.
In the absence of specific clarification, EM refers to EM sample-wise perturbation. 
All experiments are conducted on 8 GPU (NVIDIA$^\circledR$ Tesla$^\circledR$ A100$^\circledR$). 

\begin{table}[tb]
\vspace{-0.05in}
\caption{\footnotesize{CIFAR-10 test accuracy (\%) of different counter-overfitting methods. WD for weight-decay. AA for Auto-augment. RA for Rand-augment. GF for gaussian-filter.}}
\label{counter-overfitting-methods}
\vspace{-0.15in}
\begin{center}
\belowrulesep=0pt
\aboverulesep=0pt
\adjustbox{width=0.48\textwidth}{
\begin{tabular}{l|cccccc}
\toprule
Methods & Cutout~\cite{cutout} & Mixup~\cite{mixup} & Cutmix~\cite{cutmix} & AA~\cite{autoaug} \\
Test Acc. & 21.66 & 11.24 & 10.34 & 28.12 \\
\midrule
Methods & RA~\cite{randaug} & GF & Dropout~\cite{dropout} & WD~\cite{weight-decay} ($10^{-4}$) \\
Test Acc. & 27.45 & 15.95 & 25.08 & 23.51 \\
\midrule
Methods & WD ($10^{-3}$) & WD ($10^{-2}$) & $\ell_\infty$ AT & ST \\
Test Acc. & 21.99 & 21.71 & 48.16 & \textbf{82.11} \\
\bottomrule
\end{tabular}
}
\end{center}
\vskip -0.3in
\end{table}

\begin{table}[thb]
    \caption{\footnotesize{ResNet-18 test accuracies (\%) for NT and ST trained on perturbed CIFAR-10 with different protection percentages (Ratio).}}
    \label{ratio-c10}
    \belowrulesep=0pt
    \aboverulesep=0pt
    \vspace{-0.25in}
    \begin{center}
    \adjustbox{width=0.49\textwidth}{
    \begin{tabular}{l|c|cccccc}
    \toprule
    & Ratio & 0 & 0.2 & 0.4 & 0.6 & 0.8 & 1 \\
    \midrule
    \multirow{4}{*}{EM} & \normalfont{NT} & 92.41 & 93.12 & 92.04 & 90.72 & 83.90 & 19.93 \\
    &\cellcolor{gray!30} \normalfont{ST} & \cellcolor{gray!30} 82.43 & \cellcolor{gray!30} 73.70 & \cellcolor{gray!30} 71.23 & \cellcolor{gray!30} 72.18 & \cellcolor{gray!30} 72.54 & \cellcolor{gray!30} 58.63 \\
    &\cellcolor{gray!30} \normalfont{ST-CG} & \cellcolor{gray!30} 86.69 & \cellcolor{gray!30} 77.00 & \cellcolor{gray!30} 73.65 & \cellcolor{gray!30} 76.55 & \cellcolor{gray!30} 76.45 & \cellcolor{gray!30} 80.18 \\
    &\cellcolor{gray!30} \normalfont{ST-Full} & \cellcolor{gray!30} \textbf{93.79} & \cellcolor{gray!30} \textbf{93.93} & \cellcolor{gray!30} \textbf{93.70} & \cellcolor{gray!30} \textbf{92.82} & \cellcolor{gray!30} \textbf{92.86} & \cellcolor{gray!30} \textbf{87.05} \\
    \midrule
    \multirow{4}{*}{REM} & \normalfont{NT} & 92.41 & \textbf{92.15} & \textbf{91.75} & \textbf{89.29} & \textbf{84.65} & 21.86 \\
    &\cellcolor{gray!30} \normalfont{ST} & \cellcolor{gray!30} 82.43 & \cellcolor{gray!30} 82.52 & \cellcolor{gray!30} 82.61 & \cellcolor{gray!30} 82.79 & \cellcolor{gray!30} 82.65 & \cellcolor{gray!30} \textbf{82.11} \\
    &\cellcolor{gray!30} \normalfont{ST-CG} & \cellcolor{gray!30} 86.69 & \cellcolor{gray!30} 85.36 & \cellcolor{gray!30} 84.68 & \cellcolor{gray!30} 84.22 & \cellcolor{gray!30} 82.93 & \cellcolor{gray!30} 77.08 \\
    &\cellcolor{gray!30} \normalfont{ST-Full} & \cellcolor{gray!30} \textbf{93.79} & \cellcolor{gray!30} 81.04 & \cellcolor{gray!30} 78.11 & \cellcolor{gray!30} 79.97 & \cellcolor{gray!30} 78.92 & \cellcolor{gray!30} 78.86 \\
    \midrule
    \multirow{4}{*}{SP} & \normalfont{NT} & 92.41 & 93.09 & 91.50 & 90.11 & 85.19 & 13.67 \\
    &\cellcolor{gray!30} \normalfont{ST} & \cellcolor{gray!30} 82.43 & \cellcolor{gray!30} 79.22 & \cellcolor{gray!30} 70.95 & \cellcolor{gray!30} 69.86 & \cellcolor{gray!30} 64.97 & \cellcolor{gray!30} 42.73 \\
    &\cellcolor{gray!30} \normalfont{ST-CG} & \cellcolor{gray!30} 86.69 & \cellcolor{gray!30} 76.00 & \cellcolor{gray!30} 74.83 & \cellcolor{gray!30} 73.90 & \cellcolor{gray!30} 72.04 & \cellcolor{gray!30} 66.75 \\
    &\cellcolor{gray!30} \normalfont{ST-Full} & \cellcolor{gray!30} \textbf{93.79} & \cellcolor{gray!30} \textbf{93.86} & \cellcolor{gray!30} \textbf{93.32} & \cellcolor{gray!30} \textbf{93.06} & \cellcolor{gray!30} \textbf{91.97} & \cellcolor{gray!30} \textbf{79.69} \\
    \midrule
    \multirow{4}{*}{OPS} & \normalfont{NT} & 92.41 & 82.45 & 88.26 & 89.68 & 92.02 & 21.71 \\
    &\cellcolor{gray!30} \normalfont{ST} & \cellcolor{gray!30} 82.43 & \cellcolor{gray!30} 83.11 & \cellcolor{gray!30} 88.94 & \cellcolor{gray!30} 91.57 & \cellcolor{gray!30} 91.06 & \cellcolor{gray!30} \textbf{81.75} \\
    &\cellcolor{gray!30} \normalfont{ST-CG} & \cellcolor{gray!30} 86.69 & \cellcolor{gray!30} 84.09 & \cellcolor{gray!30} 89.20 & \cellcolor{gray!30} 91.02 & \cellcolor{gray!30} 87.79 & \cellcolor{gray!30} 73.09 \\
    &\cellcolor{gray!30} \normalfont{ST-Full} & \cellcolor{gray!30} \textbf{93.79} & \cellcolor{gray!30} \textbf{85.15} & \cellcolor{gray!30} \textbf{90.25} & \cellcolor{gray!30} \textbf{92.19} & \cellcolor{gray!30} \textbf{92.79} & \cellcolor{gray!30} 73.19 \\
    \bottomrule
    \end{tabular}
    }
    \end{center}
    \vskip -0.25in
\end{table}

\begin{table}[hbt]
    \caption{\footnotesize{ResNet-18 test accuracies (\%) for NT and ST trained on perturbed CIFAR-100 with different protection percentages (Ratio).}}
    \label{ratio-c100}
    \belowrulesep=0pt
    \aboverulesep=0pt
    \vspace{-0.25in}
    \begin{center}
    \adjustbox{width=0.49\textwidth}{
    \begin{tabular}{l|c|cccccc}
    \toprule
    & Ratio & 0 & 0.2 & 0.4 & 0.6 & 0.8 & 1 \\
    \midrule
    \multirow{4}{*}{EM} & \normalfont{NT} & 70.77 & \textbf{73.83} & \textbf{73.26} & 67.29 & 56.38 & 14.81 \\
    &\cellcolor{gray!30} \normalfont{ST} & \cellcolor{gray!30} 49.57 & \cellcolor{gray!30} 49.43 & \cellcolor{gray!30} 48.48 & \cellcolor{gray!30} 48.34 & \cellcolor{gray!30} 48.04 & \cellcolor{gray!30} 47.55 \\
    &\cellcolor{gray!30} \normalfont{ST-CG} & \cellcolor{gray!30} 52.31 & \cellcolor{gray!30} 51.89 & \cellcolor{gray!30} 52.12 & \cellcolor{gray!30} 50.93 & \cellcolor{gray!30} 52.00 & \cellcolor{gray!30} 51.93 \\
    &\cellcolor{gray!30} \normalfont{ST-Full} & \cellcolor{gray!30} \textbf{71.84} & \cellcolor{gray!30} 71.54 & \cellcolor{gray!30} 72.38 & \cellcolor{gray!30} \textbf{71.51} & \cellcolor{gray!30} \textbf{70.12} & \cellcolor{gray!30} \textbf{64.81} \\
    \midrule
    \multirow{4}{*}{REM} & \normalfont{NT} & 70.77 & \textbf{72.75} & \textbf{68.91} & \textbf{63.81} & \textbf{53.57} & 8.71 \\
    &\cellcolor{gray!30} \normalfont{ST} & \cellcolor{gray!30} 49.57 & \cellcolor{gray!30} 49.41 & \cellcolor{gray!30} 48.02 & \cellcolor{gray!30} 48.49 & \cellcolor{gray!30} 47.65 & \cellcolor{gray!30} 47.61 \\
    &\cellcolor{gray!30} \normalfont{ST-CG} & \cellcolor{gray!30} 52.31 & \cellcolor{gray!30} 51.62 & \cellcolor{gray!30} 50.31 & \cellcolor{gray!30} 52.73 & \cellcolor{gray!30} 51.53 & \cellcolor{gray!30} 52.94 \\
    &\cellcolor{gray!30} \normalfont{ST-Full} & \cellcolor{gray!30} \textbf{71.84} & \cellcolor{gray!30} 56.36 & \cellcolor{gray!30} 55.72 & \cellcolor{gray!30} 54.26 & \cellcolor{gray!30} \textbf{55.32} & \cellcolor{gray!30} 54.19 \\
    \midrule
    \multirow{4}{*}{SP} & \normalfont{NT} & 70.77 & \textbf{73.68} & \textbf{71.37} & 67.55 & 57.88 & 9.54 \\
    &\cellcolor{gray!30} \normalfont{ST} & \cellcolor{gray!30} 49.57 & \cellcolor{gray!30} 47.07 & \cellcolor{gray!30} 43.77 & \cellcolor{gray!30} 40.84 & \cellcolor{gray!30} 37.50 & \cellcolor{gray!30} 32.54 \\
    &\cellcolor{gray!30} \normalfont{ST-CG} & \cellcolor{gray!30} 86.69 & \cellcolor{gray!30} 76.00 & \cellcolor{gray!30} 74.83 & \cellcolor{gray!30} 73.90 & \cellcolor{gray!30} 72.04 & \cellcolor{gray!30} 66.75 \\
    &\cellcolor{gray!30} \normalfont{ST-Full} & \cellcolor{gray!30} \textbf{71.84} & \cellcolor{gray!30} 71.97 & \cellcolor{gray!30} 70.06 & \cellcolor{gray!30} \textbf{71.48} & \cellcolor{gray!30} \textbf{68.98} & \cellcolor{gray!30} \textbf{52.46} \\
    \midrule
    \multirow{4}{*}{OPS} & \normalfont{NT} & 70.77 & 41.09 & 58.93 & 64.94 & \textbf{69.93} & 13.93 \\
    &\cellcolor{gray!30} \normalfont{ST} & \cellcolor{gray!30} 49.57 & \cellcolor{gray!30} 42.15 & \cellcolor{gray!30} 59.06 & \cellcolor{gray!30} 60.19 & \cellcolor{gray!30} 53.12 & \cellcolor{gray!30} 29.98 \\
    &\cellcolor{gray!30} \normalfont{ST-CG} & \cellcolor{gray!30} 52.31 & \cellcolor{gray!30} 50.18 & \cellcolor{gray!30} 59.75 & \cellcolor{gray!30} 66.18 & \cellcolor{gray!30} 57.57 & \cellcolor{gray!30} \textbf{47.73} \\
    &\cellcolor{gray!30} \normalfont{ST-Full} & \cellcolor{gray!30} \textbf{71.84} & \cellcolor{gray!30} \textbf{51.47} & \cellcolor{gray!30} \textbf{60.93} & \cellcolor{gray!30} \textbf{67.25} & \cellcolor{gray!30} 68.43 & \cellcolor{gray!30} 46.73 \\
    \bottomrule
    \end{tabular}
    }
    \end{center}
    \vskip -0.3in
\end{table}

\subsection{Learning Effectiveness.}
\subsubsection{Experimental Result}
Tab.~\ref{r_CIFAR-10} shows the clean test accuracy of models on CIFAR-10 after being protected with different methods. As a pioneering method, EM perturbations have limited data protection capabilities, while methods on improving it like REM and OPS show that the natural training method cannot work well on perturbed data, most of which are less than 30\% on CIFAR-10. Adversarial training (AT) works well on EM perturbations. However, AT is still less effective compared with our ST training pipeline. Especially, ST-Full trained ResNet-50 gets 92.84\% on EM perturbed CIFAR-10. Meanwhile, the ST training pipeline works effectively on clean data as well as normal training indicating our method can be an add-on to the NT whether the training data is protected or not. 

Tab.~\ref{r_CIFAR-100} shows the clean test accuracy of models on CIFAR-100 after being trained with different methods on the perturbed training data. 
As a pioneering method, EM perturbations have limited data protection capabilities, while methods on improving it like REM and OPS show that the natural training method cannot work well on perturbed data, most of which are less than 17\% on CIFAR-100. Adversarial training (AT) works well on EM and SP perturbations. However, AT is still trivial compared with our ST training pipeline. Especially, ST-Full trained ResNet-50 gets 71.94\% on EM perturbed CIFAR-100. Meanwhile, the ST training pipeline works effectively on clean data as well as normal training.

Furthermore, to confirm the effectiveness of high-resolution images, we implement the ST training pipeline on ImageNet-mini. As shown in Tab.~\ref{r-imagnet-mini}, ST works well on four perturbed data (especially, ST improves ResNet-18 from 11.10\% to 75.59\% on SP and DenseNet-121 from 6.58\% to 65.26\% on EM). CG augmentation supplements the performance of ST, which ST-CG achieves a good boost to ST. Meanwhile, ST-Full works perfectly on REM and SP perturbation. However, it has a decreased accuracy on EM with ResNet-18. This is because the protection of perturbation on high-resolution images is much stronger. For example, NT trained ResNet-18 on EM perturbations can achieve 19.93\% accuracy in Tab.~\ref{r_CIFAR-10} while it decreases to 14.81\% for  ImageNet-mini in Tab.~\ref{r-imagnet-mini}.

\begin{figure}[bht]
\begin{center}
	\includegraphics[width=\linewidth]{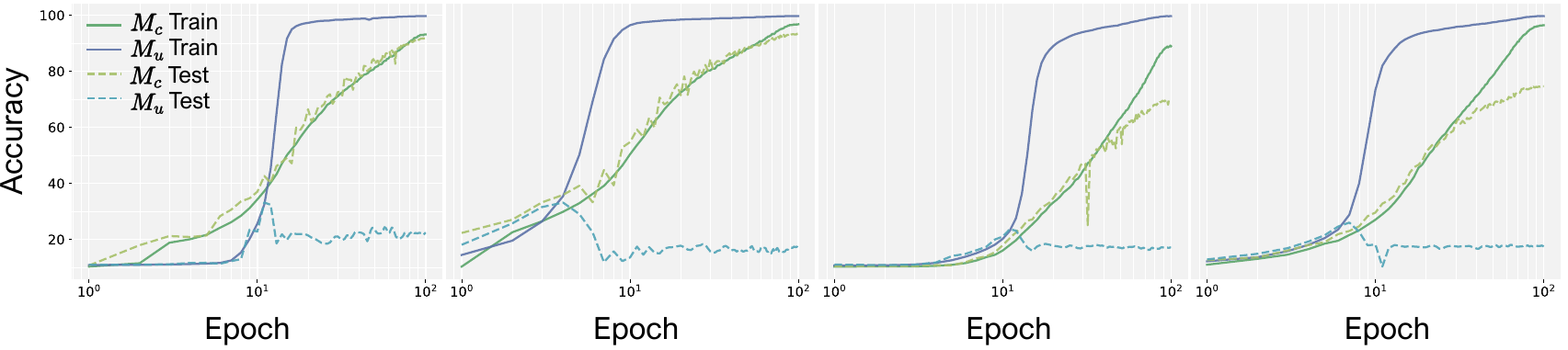}
    \vspace{-0.25in}
\caption{\footnotesize{From left to right, the sequence is as follows: the training and test accuracy of a ResNet-50 trained on clean CIFAR-10 ($\bm{\theta}_c$) and REM perturbed CIFAR-10 ($\bm{\theta}_u$), the training and test accuracy of a WRN34-10 trained on clean CIFAR-10 ($\bm{\theta}_c$) and perturbed CIFAR-10 ($\bm{\theta}_u$), the training and test accuracy of a ResNet-50 trained on clean CIFAR-100 ($\bm{\theta}_c$) and REM perturbed CIFAR-100 ($\bm{\theta}_u$), the training and test accuracy of a WRN34-10 trained on clean CIFAR-100 ($\bm{\theta}_c$) and perturbed CIFAR-100 ($\bm{\theta}_u$).}}
\label{app-nt-clean-poison}
\end{center}
\vskip -0.2in
\end{figure}

\begin{figure}[tbh]
\begin{center}
	\includegraphics[width=\linewidth]{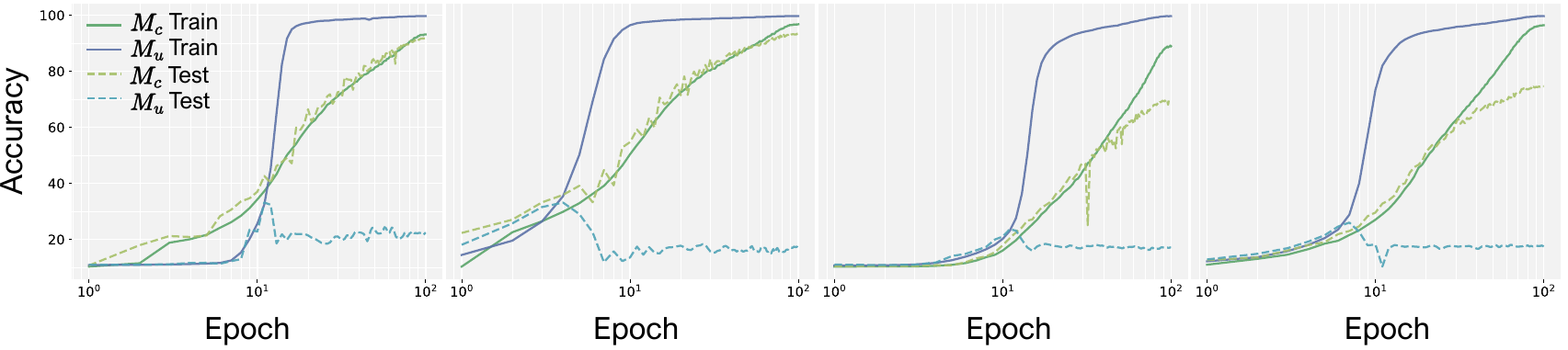}
    \vspace{-0.25in}
\caption{\footnotesize{From left to right, the sequence is as follows: the training and test accuracy of $\bm{\theta}_u^S$ (green) and $\bm{\theta}_u^D$ (blue) with VGG-16 on REM perturbed CIFAR-10, the training and test accuracy of $\bm{\theta}_u^S$ (green) and $\bm{\theta}_u^D$ (blue) with WRN34-10 on REM perturbed CIFAR-10, the training and test accuracy of $\bm{\theta}_u^S$ (green) and $\bm{\theta}_u^D$ (blue) with VGG-16 on REM perturbed CIFAR-100, the training and test accuracy of $\bm{\theta}_u^S$ (green) and $\bm{\theta}_u^D$ (blue) with VGG-16 on REM perturbed CIFAR-100.}}
\label{app-replace-sd}
\end{center}
\vskip -0.2in
\end{figure}

\subsubsection{Comparison with general counter-overfitting methods.}
\label{c-overfitting}
To investigate whether counter-overfitting methods can correctly guide the model to escape from unlearnable perturbation feature learning, we tested several counter-overfitting methods. As shown in Tab.~\ref{counter-overfitting-methods}, we naturally trained a ResNet-18 on REM-perturbed CIFAR-10. 
The data augmentations, including Cutout, Mixup, Cutmix, Auto-augment, Rand-augment, and Gaussian filter, ineffectively resist unlearnable examples. 
Then, we implement dropout (0.5 probability) and weight-decay (WD) from $10^{-4}$ to $10^{-2}$, both of which cannot defeat unlearnable examples. Finally, we use adversarial training (AT) with $\ell_\infty\leq\frac{4}{255}$. 
ST emerges as the most effective strategy for overcoming unlearnable examples, whereas traditional counter-overfitting methods fail to systematically handle training processes.

\subsubsection{Different protection percentages.}
A more realistic learning scenario is that only a part of the data is protected by the defensive noise and the others are clean. We used various unlearnable methods to perturb CIFAR-10 and CIFAR-100 with different mixing ratios. The protection percentage (ratio in Tab.~\ref{ratio-c10} and Tab~\ref{ratio-c100}) represents the proportion of unlearnable examples to all samples. As shown in Tab~\ref{ratio-c10} and Tab~\ref{ratio-c100}, ST-Full has exceptional performance on all protection percentages (especially on protection percentage 1.0), which reflects the reliability of our method on mixed samples. 

\begin{figure}[tb]
\begin{center}
	\centering
	\includegraphics[width=0.85\columnwidth,height=0.567\columnwidth]{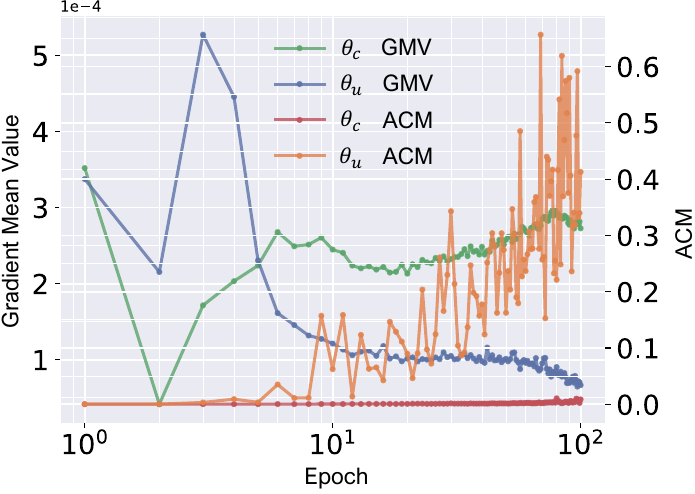}
        \vspace{-0.1in}
\caption{\footnotesize{(c) The mean value of gradient (GMV) of a shallow layer and \textit{ACM} at each epoch when trained on clean data $\bm{\theta}_c$ and unlearnable data $\bm{\theta}_u$.}}
\label{plot_grad_acm}
\end{center}
\vskip -0.2in
\end{figure}

\subsection{Analysis and Ablation Study}
\label{analysis}
\subsubsection{Observations for insights}
\label{Observations_for_insights}

\begin{figure}[tb]
\begin{center}
	\centering
	\begin{minipage}{\linewidth}
		\centering
		\includegraphics[width=0.95\linewidth]{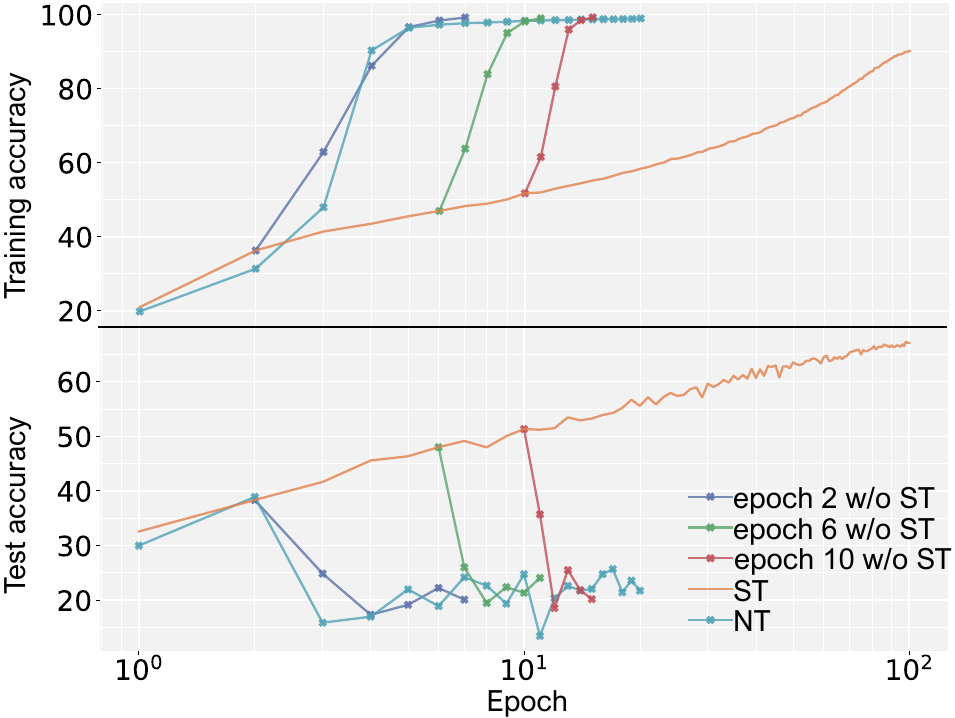}
	\end{minipage}
\vspace{-0.1in}
\caption{\footnotesize{The training and test accuracy of a RensNet-18 ST trained on REM perturbed data (orange), and the accuracy of not implementing ST (blue, green, red). The model will drop to an unlearnable perturbation feature learning trap without ST. NT (same as $\bm{\theta}_u$ in Fig.~\ref{curves}) is a ResNet-18 normal trained on REM perturbed data (light blue)}}
\label{sft-accuray}
\end{center}
\vspace{-0.3in}
\end{figure}

\begin{figure*}[t]
\begin{center}
	\includegraphics[width=0.95\linewidth]{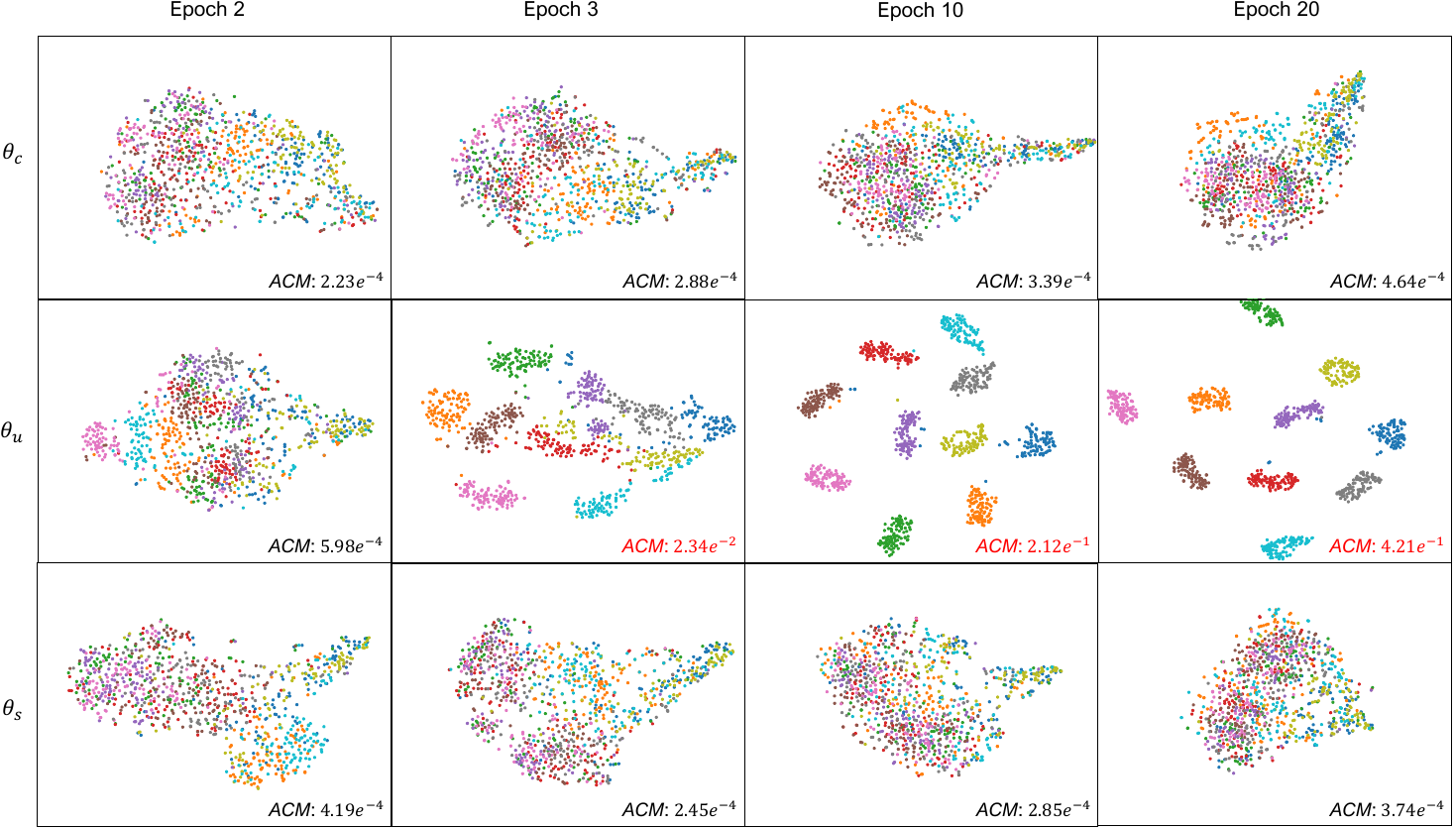}
    \vspace{-0.1in}
\caption{\footnotesize{The fourth residual block layer activation t-SNE results and \textit{ACM} (~Eq.\eqref{eq:acm}) of various models in different epochs. $\bm{\theta}_c$ is a model naturally trained on clean data. $\bm{\theta}_u$ is a model naturally trained on unlearnable data. $\bm{\theta}_s$ is a model ST trained on unlearnable data.}}
\label{layer-2-tSNE}
\end{center}
\vskip -0.3in
\end{figure*}

\begin{figure}[tbh]
\begin{center}
	\includegraphics[width=\linewidth]{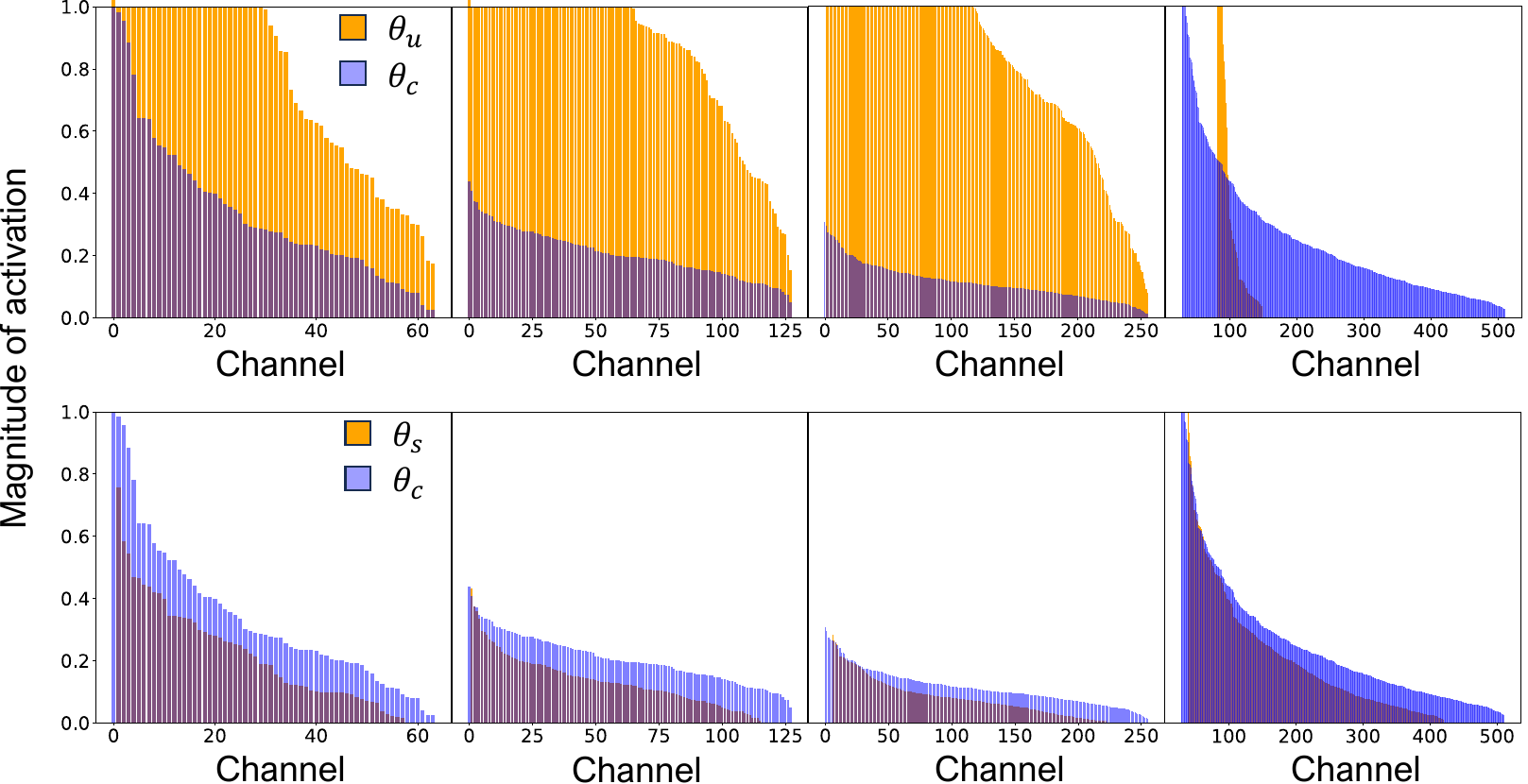}
    \vspace{-0.2in}
\caption{\footnotesize{The first row of images are the channel-wise mean value of output activation from the shallow layers to the deep layers of model $\bm{\theta}_c$ and $\bm{\theta}_u$, while the second row of images is the channel-wise mean value of output activation from the shallow layers to the deep layers of model $\bm{\theta}_c$ and $\bm{\theta}_s$. 
The first, second, and third columns correspond to the output activation distribution of the 2nd, 4th, and 6th residual block layers, respectively. The fourth column is the output activation distribution of the residual block layer before the full connection layer.
}}
\label{app-channel-activation}
\end{center}
\vskip -0.3in
\end{figure}

\begin{figure*}[t]
\begin{center}
	\includegraphics[width=0.95\linewidth]{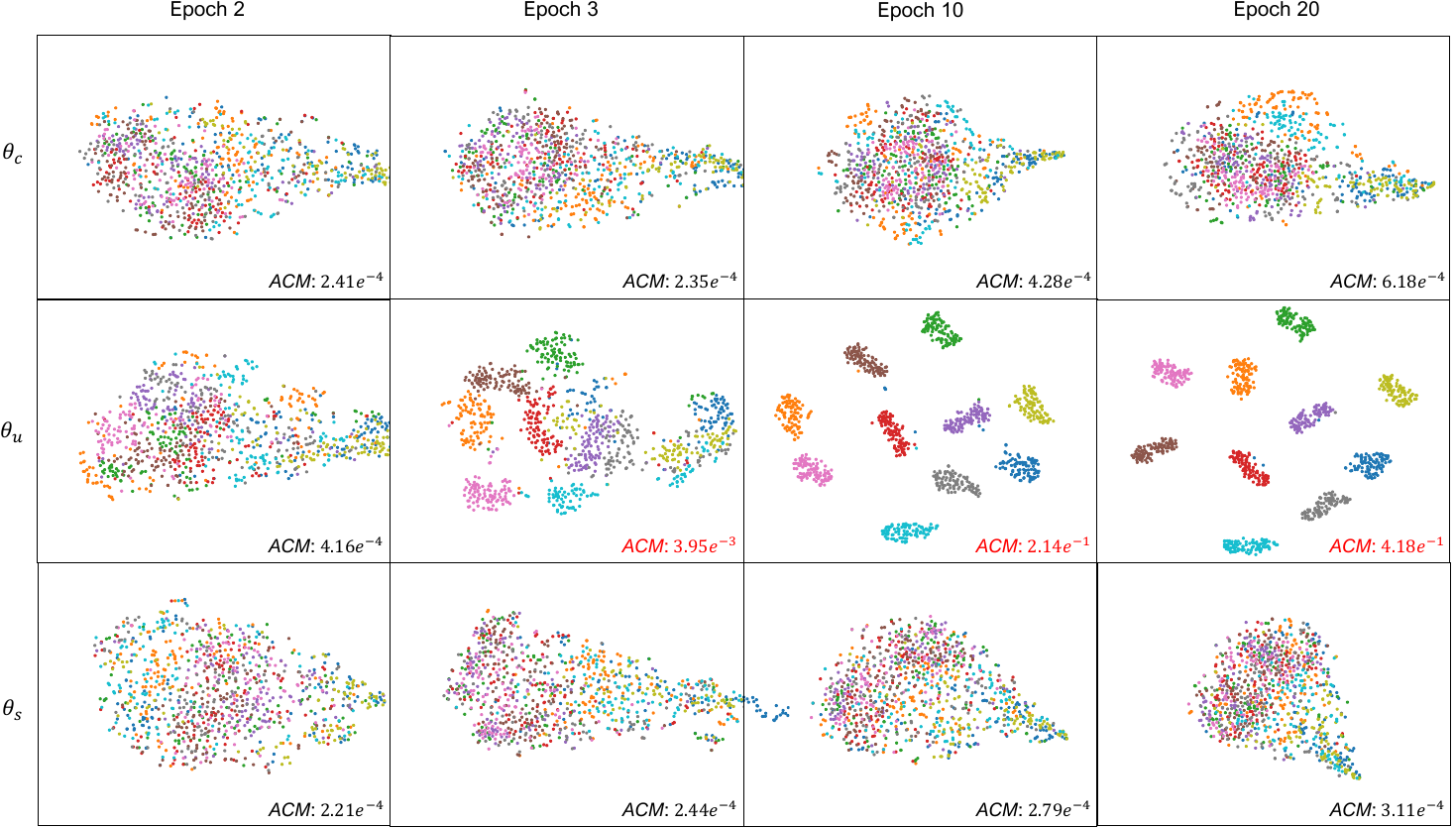}
    \vspace{-0.1in}
\caption{\footnotesize{The sixth residual block layer activation t-SNE results and \textit{ACM} (~Eq.\eqref{eq:acm}) of various models in different epochs. $\bm{\theta}_c$ is a model naturally trained on clean data. $\bm{\theta}_u$ is a model naturally trained on unlearnable data. $\bm{\theta}_s$ is a model ST trained on unlearnable data.}}
\label{layer-3-tSNE}
\end{center}
\vskip -0.3in
\end{figure*}

We provide additional experiments to support our observations, including various architectures and datasets.

\textbf{Observations for Insight 1.}
We also show the learning process of a ResNet-50 and a WRN34-10 separately on unlearnable CIFAR-10 and CIFAR-100 in Fig.~\ref{app-nt-clean-poison}. Both the training accuracy and test accuracy increase at the first several epochs, indicating that the model learned the correct image features. However, as training goes on, the training accuracy increases sharply, while the test accuracy significantly decreases, indicating the model is trapped in the perturbation feature learning. These observations provide direct evidence for our insight 1.

\textbf{Observations for Insight 2.}
We also show the learning process of a ResNet-50 and a WRN34-10 separately on unlearnable CIFAR-10 and CIFAR-100. As shown in Fig.~\ref{app-replace-sd}, $\bm{\theta}_u^S$ performs better test data, which proves that shallow layers are capable of learning the correct features at the beginning of the training (like $\bm{\theta}_u^S$). However, when shallow layers are trapped in unlearnable perturbation feature learning, even if deep layers are in the right state (like $\bm{\theta}_u^D$), shallow layers will pass the wrong activation and corrupt deep layers to a wrong learning way. These observations provide direct evidence for our insight 2.

\subsubsection{\textit{ACM} Analysis}
\label{ACM_analysis}
We natural trained a ResNet-18 on clean data ($\bm{\theta}_c$) and unlearnable data ($\bm{\theta}_u$), and compare them with a ResNet-18 ST trained on unlearnable data ($\bm{\theta}_s$). Then we drew the t-SNE cluster results of the output activation of the fourth residual block layer (Fig.~\ref{layer-2-tSNE}) and the sixth residual block layer (Fig.~\ref{layer-3-tSNE}) of ResNet-18. The results show that when a raw model is trained on unlearnable data, the activation cluster disorder at early epochs. As training goes by, the model begins to cluster well, which is a phenomenon of unlearnable perturbation feature learning. The visualization results of t-SNE also reflect the wrong learning process during training.

The \textit{ACM} of the model trained on clean data ($\bm{\theta}_c$) and unlearnable data ($\bm{\theta}_u$) at different training epochs is shown in Fig.~\ref{plot_grad_acm}. The \textit{ACM} of the $\bm{\theta}_u$ is low at the beginning, then it goes to a high level, which is consistent with our previous observation in t-SNE cluster results that the inter-class distance becomes larger and intra-class distance becomes smaller. Meanwhile, The gradient mean value (GMV) of the model trained on unlearnable data falls to a low level rapidly, which illustrates that shallow layers are trapped in unlearnable perturbation feature learning and hardly learn image semantic features at late epochs compared to the model trained on clean data.

\subsubsection{ST helps to learn image semantic features}
\label{help_aviod_overfitting}
We further illustrate how the ST framework helps to avoid unlearnable perturbation feature learning. We first train a model on REM perturbed data. When this model tends to learn unlearnable perturbation feature ($\textit{ACM}>\gamma$), we either keep ST training (orange lines in Fig.~\ref{sft-accuray}) or shift to natural training (blue, green, red lines in Fig.~\ref{sft-accuray}).
It demonstrates that without our learning rate adjustment algorithm, the model will fall back to unlearnable perturbation feature learning as natural training, which shows the necessity of learning rate adjustment. 

To verify the effectiveness of ST, we analyze the channel-wise activation of different layers. We natural trained a ResNet-18 model separately on clean data ($\bm{\theta}_c$) and REM perturbed data ($\bm{\theta}_u$). Then we ST trained a ResNet-18 on REM perturbed data ($\bm{\theta}_s$). We analyze the channel-wise mean value of output activation from the shallow layers to the deep layers of model $\bm{\theta}_c$, $\bm{\theta}_u$ and $\bm{\theta}_s$, then feed them with clean data. As shown in Fig.~\ref{app-channel-activation}, the activation distribution of $\bm{\theta}_s$ is similar with $\bm{\theta}_c$, while the activation distribution of $\bm{\theta}_u$ is largely different with $\bm{\theta}_c$ and $\bm{\theta}_s$. In this case, ST resists unlearnable perturbation feature learning and correctly leads the model to learn image semantic features.

\subsubsection{Hyperparameter analysis.}
\label{hyperparameter_analysis}
\begin{figure}[tbh]
\begin{center}
	\includegraphics[width=\linewidth]{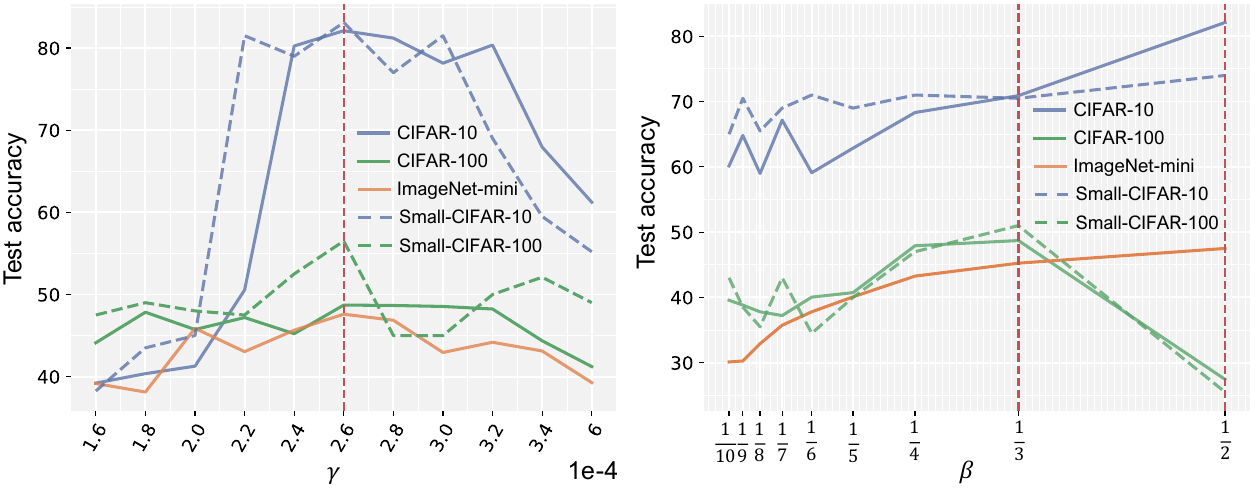}
    \vspace{-0.3in}
\caption{\footnotesize{\textit{Left}: The test accuracy of a RensNet-18 ST-CG trained on REM perturbed data. \textit{Right}: The test accuracy of a ResNet-18 ST-CG trained on REM perturbed data.
}}
\label{gamma_alpha}
\end{center}
\vskip -0.3in
\end{figure}

We analyzed the sensitivity of hyperparameter $\gamma$ and $\beta$ shown in Fig.~\ref{gamma_alpha}. $2.6 \times 10^{-4}$ is a suitable value for $\gamma$. Meanwhile, $\beta=50\%$ is a suitable value for CIFAR-10, ImageNet-mini, and $\beta=33.33\%$ is suitable for CIFAR-100. Especially, we analyzed the sensitivity of hyperparameter $\gamma$ and $\beta$ on a validation set containing only 20 clean samples of the test dataset, and results show that 20 clean samples are sufficient for picking a suitable hyperparameter. In this case, while deploying our method, the engineer could collect a small validation set (only 20 samples) and pick suitable hyperparameters.

\subsubsection{Adaptive Attack}
\label{adaptive_attack}
\textbf{\emph{Our method works well even if CG becomes a white box setting while generating perturbation.}}
We propose composite data augmentation strategies to improve the learning effectiveness of unlearnable examples. In this subsection, we analyze that the CG-transformed image data lowers the chance of unlearnable perturbation feature learning. Such composite data augmentation can work effectively even if the augmentation strategy is white-box during generating unlearnable examples because it is hard to build effective unlearnable examples with more complex perturbation feature space. 
Suppose we have a dataset consisting of original clean examples $\mathcal{D}_{clean}=\{\left(\bm{x}_i,{y}_i\right)\}_{i=1}^{n}$ drawn from a distribution $\mathcal{S}$, where $\bm{x}_i\in \mathcal{X}$ is an input image and ${y}_i\in\mathcal{Y}$ is the associated label. We assume that the unauthorized parties will use the published training dataset to train a classifier $f_{\bm{\theta}}:\mathcal{X}\rightarrow\mathcal{Y}$ with parameter $\bm{\theta}$. 
We implement augmentation CG while generating EM and EM class-wise perturbations, and we denote these perturbations by EM-T and EM-Class-T which generate as follows:
\begin{equation}\label{EM-T}
    \arg\min_{\bm{\theta}}\mathbb{E}_{(\bm{x}_i,{y_i)}}\left[\min_{\bm{\delta}_i}\mathcal{L}\left(f_{\bm{\theta}}(\mathcal{T}(\bm{x_i}+\bm{\delta}_i)),{y_i}\right)\right]
\end{equation}
where $\Vert\bm{\delta}_{i}\Vert_{\infty}<\epsilon_u$, $\mathcal{T}(\cdot)$ is the augmentation CG and $\mathcal{L}$ is the loss function. Meanwhile, We implement augmentation CG while generating REM perturbations, and we denote these perturbations by REM-T which generate as follows:
\begin{equation}\label{REM-T}
    \arg\min_{\bm{\theta}}\mathbb{E}_{(\bm{x_i},{y_i)}}\left[\min_{\bm{\delta}_i}\max_{\bm{\sigma}_i}\mathcal{L}\left(f_{\bm{\theta}}(\mathcal{T}(\bm{x_i}+\bm{\delta}_i+\bm{\sigma}_i)),{y_i}\right)\right]
\end{equation}
where $\Vert\bm{\sigma}_{i}\Vert_{\infty}<\epsilon_a$, $\Vert\bm{\delta}_{i}\Vert_{\infty}<\epsilon_u$, $\mathcal{T}(\cdot)$ is the augmentation CG and $\mathcal{L}$ is the loss function. Following Eq.~\ref{EM-T} and Eq.~\ref{REM-T}, we set CG as a white box set. As shown in Tab.~\ref{white-box}, our method works well with EM-T, EM-Class-T, and REM-T.

\textbf{\emph{Our method works well even if ST becomes a white box setting while generating perturbation.}} We implement ST while generating EM, EM-Class, and REM perturbations. We denote these perturbations by EM-A, EM-Class-A, and REM-A. Meanwhile, we set augmentation CG and ST as white box settings while generating EM, EM-Class, and REM perturbations, and we denote these perturbations by EM-AT, EM-Class-AT, and REM-AT. As shown in Tab.~\ref{white-box}, such white-box settings will not contribute to data protection capabilities indicating the effectiveness of our method.

\begin{table}[htb]
    \vspace{-0.1in}
    \caption{\footnotesize{CIFAR-10 clean test accuracies (\%) for natural training (NT), natural training with augment CG (NT-CG) and ST in different versions (ours) with ResNet18 on clean samples and various unlearnable examples. EM is for sample-wise EM perturbations and EM-Class is for class-wise EM perturbations.}}
    \label{white-box}
    \belowrulesep=0pt
    \aboverulesep=0pt
    \vspace{-0.15in}
    \begin{center}
    \adjustbox{width=0.48\textwidth}{
        \begin{tabular}{l|cccccc}
            \toprule
            Method & \normalfont{NT} & \normalfont{NT-CG} & \normalfont{ST} & \normalfont{ST-CG} & \normalfont{ST-Full} \\
            \midrule
            \normalfont{Clean} & 92.41 & 93.20 & \cellcolor{gray!30} 82.43 & \cellcolor{gray!30} 86.62 & \cellcolor{gray!30} \textbf{93.79}\\
            \midrule
            \normalfont{EM} & 19.93 & 60.03 & \cellcolor{gray!30} 58.63 & \cellcolor{gray!30} 78.44 & \cellcolor{gray!30} \textbf{87.05}\\
            \normalfont{EM-T} &23.19 & 80.30 & \cellcolor{gray!30} 68.23 & \cellcolor{gray!30} 78.47 & \cellcolor{gray!30} \textbf{81.19} \\
            \normalfont{EM-A} & 91.57 & 92.86 & \cellcolor{gray!30} 69.33 & \cellcolor{gray!30} 69.79 & \cellcolor{gray!30} \textbf{93.48} \\
            \normalfont{EM-AT} & 93.29 & 93.01 & \cellcolor{gray!30} 71.18 & \cellcolor{gray!30} 76.13 & \cellcolor{gray!30} \textbf{93.54} \\
            \midrule
            \normalfont{EM-Class} & 9.39 & 34.22 & \cellcolor{gray!30} 46.29 & \cellcolor{gray!30} 71.08 & \cellcolor{gray!30} \textbf{80.04} \\
            \normalfont{EM-Class-T} & 17.05 & 77.7 & \cellcolor{gray!30} 55.96 & \cellcolor{gray!30} 65.72 & \cellcolor{gray!30} \textbf{76.08}\\
            \normalfont{EM-Class-A} & 92.80 & 93.12 & \cellcolor{gray!30} 69.66 & \cellcolor{gray!30} 73.16 & \cellcolor{gray!30} \textbf{93.63} \\
            \normalfont{EM-Class-AT} & 76.43 & \textbf{92.81} & \cellcolor{gray!30} 68.15 & \cellcolor{gray!30} 73.79 & \cellcolor{gray!30} 85.87\\
            \midrule
            \normalfont{REM} & 21.86 & 55.62 & \cellcolor{gray!30} \textbf{82.11} & \cellcolor{gray!30} 77.08 & \cellcolor{gray!30} 78.86 \\
            \normalfont{REM-T} & 61.03 & 72.36 & \cellcolor{gray!30} 65.06 & \cellcolor{gray!30} 68.15 & \cellcolor{gray!30} \textbf{78.34} \\
            \normalfont{REM-A} & 35.22 & 81.84 & \cellcolor{gray!30} 54.59 & \cellcolor{gray!30} 64.47 & \cellcolor{gray!30} \textbf{84.93} \\
            \normalfont{REM-AT} & 60.13 & 79.93 & \cellcolor{gray!30} 55.52 & \cellcolor{gray!30} 65.72 & \cellcolor{gray!30} \textbf{81.81} \\
            \bottomrule
    \end{tabular}
    }
    \end{center}
    \vskip -0.1in
\end{table}

\subsubsection{Data Augmentation. }
\label{whycg}
To verify the effectiveness of color-jitter and gray-scale (CG). Color jitter randomly changes an image's brightness, contrast, saturation, and hue. Gray-scale randomly converts an image to grayscale. We have conducted experiments on REM perturbed CIFAR-10 using various data augmentations on ResNet-18, and the results indicate that not all data augmentations can break unlearnable perturbations (Table~\ref{r-t-2}). In this case, we utilized CG as an auxiliary means to enhance the performance of ST.

\begin{table}[htb]
\vspace{-0.1in}
\caption{\footnotesize{Test accuracy (\%) of normal trained ResNet-18 on REM perturbed CIFAR-10 with various augmentations.}}
\label{r-t-2}
\vspace{-0.2in}
\belowrulesep=0pt
\aboverulesep=0pt
\begin{center}
\adjustbox{width=0.48\textwidth}{
\begin{tabular}{l|ccccc}
\toprule
Methods & Mixup & Cutmix & Gamma & Brightness & Solarize \\
Test Acc. & 11.24 & 10.34 & 17.76 & 20.46 & 22.92 \\
\midrule
Methods & Hue & Grayscale & Colorjitter & NT-CG & ST \\
Test Acc. & 18.71 & 45.62 & 37.21 & 55.62 & \textbf{82.11} \\
\bottomrule
\end{tabular}
}
\end{center}
\vspace{-0.2in}
\end{table}

\subsubsection{Comparison with defensive measures against data poisoning attacks. }
\label{poison_attack_defense}
Currently, there is a lack of methods to counteract unlearnable examples that can be compared with our approach ST, while data poison attacks share the same principle with unlearnable examples. 
In this case, we verified the efficiency of different defensive measures against data poisoning attacks~\cite{EPI,friends} with a ResNet-18 on EM, REM, SP perturbed CIFAR-10 in Tab.~\ref{counter-poison-attack}. 
The results indicate that the defensive measures against data poisoning attacks are ineffective against unlearnable examples, as the experimental outcomes are nearly equivalent to the results of random guessing. In contrast, ST effectively overcomes unlearnable examples. 
This is because although data poisoning attacks share the same principles as unlearnable examples, they serve different application scenarios. Therefore, defense methods against data poisoning attacks are ineffective for unlearnable examples.

\begin{table}[htb]
\vspace{-0.1in}
\caption{\footnotesize{CIFAR-10 test accuracy (\%) of different defensive measures against data poisoning attacks. }}
\label{counter-poison-attack}
\vspace{-0.25in}
\begin{center}
\belowrulesep=0pt
\aboverulesep=0pt
\adjustbox{width=0.49\textwidth}{
\begin{tabular}{l|cccccc}
\toprule
Methods & EPI~\cite{EPI} & FrieNDs~\cite{friends} & \cellcolor{gray!30} ST & \cellcolor{gray!30} ST-CG & \cellcolor{gray!30} ST-Full \\
\midrule
EM & 9.98 & 11.76 & \cellcolor{gray!30} 58.63 & \cellcolor{gray!30} 80.18 & \cellcolor{gray!30} \textbf{87.05} \\
\midrule
REM & 11.21 & 10.58 & \cellcolor{gray!30} \textbf{82.11} & \cellcolor{gray!30} 77.08 & \cellcolor{gray!30} 78.86 \\
\midrule
SP & 10.54 & 10.27 & \cellcolor{gray!30} 42.73 & \cellcolor{gray!30} 66.75 & \cellcolor{gray!30} \textbf{79.69}\\
\bottomrule
\end{tabular}
}
\end{center}
\vskip -0.25in
\end{table}

\section{Conclusion}
In this paper, we study the mechanism of the data protection method, unlearnable examples that typically cause models to ignore correct semantic features and learn incorrect perturbation features instead. We observed that unlearnable examples mislead models into a trap of unlearnable perturbation learning. We propose the Activation Cluster Measurement (\textit{ACM}) to quantify the unlearnable perturbation learning degree of a model. Based on that, we propose progressive staged training (ST), a novel stage training framework to gradually slow down the learning process from shallow layers to deep layers and defeat unlearnable examples. Our ST training pipeline breaks all existing state-of-the-art methods and provides a reliable evaluation benchmark for further studies on unlearnable examples.

\bibliographystyle{IEEEtran}
\bibliography{IEEEabrv,main}

\vspace{-0.5in}

\vfill

\end{document}